\pdfoutput=1

\documentclass[11pt]{article}

\usepackage{acl}

\usepackage{times}
\usepackage{latexsym}

\usepackage[T1]{fontenc}

\usepackage[utf8]{inputenc}

\usepackage{microtype}

\usepackage{amsmath}
\usepackage{graphicx}
\usepackage{inconsolata}
\usepackage{booktabs}       
\usepackage{amsfonts}       
\usepackage{nicefrac}       
\usepackage{xcolor}         
\usepackage{float}
\usepackage{multirow,diagbox,enumitem}
\usepackage{arydshln}
\usepackage{tabularx}
\usepackage{algorithm}
\usepackage{algorithmic}
\usepackage{cleveref}

\newcommand{\vpara}[1]{\vspace{0.05cm}\noindent\textbf{#1}}

\newcommand{\defense}[0]{{PBA }}
\newcommand{\attack}[0]{{NAW }}

\title{Navigation as Attackers Wish? Towards Building Robust Embodied Agents under Federated Learning}

\begin{document}

\def\eg{\emph{e.g}.} \def\Eg{\emph{E.g}.}
\def\ie{\emph{i.e}.} \def\Ie{\emph{I.e}.}
\def\cf{\emph{c.f}.} \def\Cf{\emph{C.f}.}
\def\etc{\emph{etc}.} \def\vs{\emph{vs}.}
\def\wrt{w.r.t.} \def\dof{d.o.f.}
\def\etal{\emph{et al}.}

\author{Yunchao Zhang\thanks{This work is done while Yunchao is interning at UCSC}, \quad Zonglin Di, \quad Kaiwen Zhou, \quad Cihang Xie, \quad Xin Eric Wang\\
University of California, Santa Cruz\\
{\tt\small \{yzhan885,zdi,kzhou85,cixie,xwang366\}@ucsc.edu}
}
\maketitle
\begin{abstract}

Federated embodied agent learning~\citep{zhou2022fedvln} protects the data privacy of individual visual environments by keeping data locally at each client (the individual environment) during training.
However, since the local data is inaccessible to the server under federated learning, attackers may easily poison the training data  of the local client to build a backdoor in the agent without notice. Deploying such an agent raises the risk of potential harm to humans, as the attackers may easily navigate and control the agent as they wish via the backdoor. 
Towards Byzantine-robust federated embodied agent learning, in this paper, we study the attack and defense for the task of vision-and-language navigation (VLN), where the agent is required to follow natural language instructions to navigate indoor environments.
First, we introduce a simple but effective attack strategy, Navigation as Wish (NAW), in which the malicious client manipulates local trajectory data to implant a backdoor into the global model. Results on two VLN datasets (R2R~\citep{r2r} and RxR~\citep{rxr}) show that NAW can easily navigate the deployed VLN agent regardless of the language instruction, without affecting its performance on normal test sets. 
Then, we propose a new Prompt-Based Aggregation (PBA) to defend against the NAW attack in federated VLN, which provides the server with a ``prompt'' of the vision-and-language alignment variance between the benign and malicious clients so that they can be distinguished during training. 
We validate the effectiveness of the \defense method on protecting the global model from the \attack attack, which outperforms other state-of-the-art defense methods by a large margin in the defense metrics on R2R and RxR.
\end{abstract}

\section{Introduction}
\begin{figure}[!t]
    \includegraphics[width=\columnwidth]{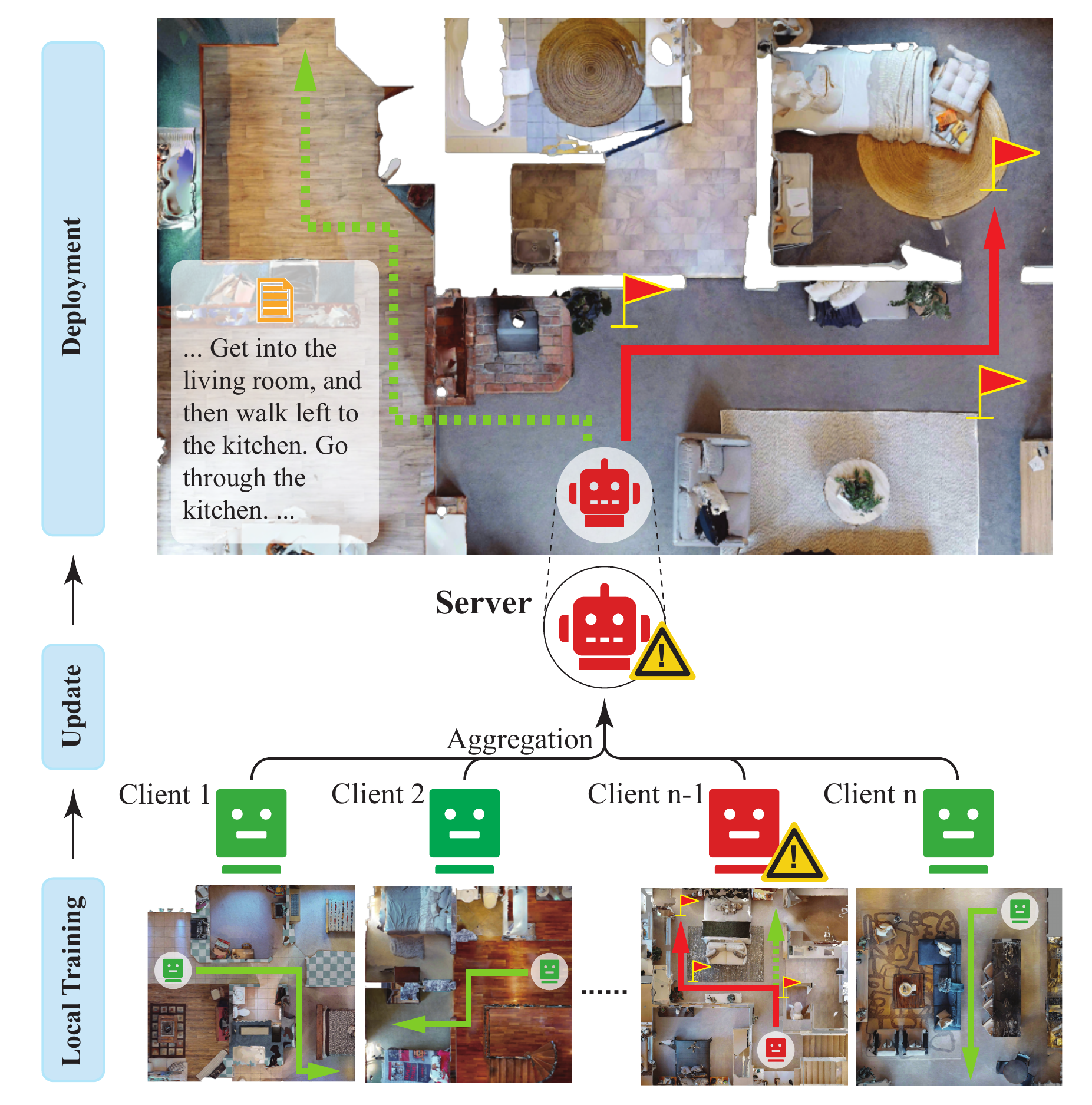}
    \caption{Illustration for the targeted backdoor attack in federated vision-and-language navigation. The green clients refer to the benign clients with ground-truth training data, while the red client refers to the malicious client (attacker) with poisoned training data. The ref flag added in the view is the trigger from the attacker. 
    With the targeted attack, the agent will miss the correct route (green line) and turn to the expected route as the attacker wishes without following the language instruction.
    }
    \vspace{-3ex}
    \label{intro}
\end{figure}

Building embodied agents that can understand the environment and perform real-world tasks following human instructions has been a long-standing goal of the AI research community. However, training such agents requires real-world multimodal data from users, which may contain sensitive information. Federated learning~\citep{fedavg,zhou2022fedvln} (FL) has been used to protect data privacy in embodied agent learning on the task of vision-and-language navigation (VLN)~\citep{r2r}, in which an agent is required to navigate to a target location following language instruction. 
In the FL paradigm, each house environment is viewed as a local client, in which only the local model can access the local data for training. 
The clients will upload their local models to the server periodically in FL, but there is no data communication between the server and the clients, so the privacy of the local data of individual environments is better preserved.

However, due to the lack of transparency in the local training process, federated learning has been shown to be vulnerable to attack methods~\citep{bhagoji2019analyzing,Lyu2020PrivacyAR}. Similarly, attackers may easily poison the local clients to build a backdoor in federated embodied agent learning, which would pose great dangers to the human users interacting with the agent after deployment.
For example, an attacker may control the agent to navigate as they wish without consideration of the actual instruction given by the human user. 
This paper studies the unique attack and defense problems in Federated Vision-and-Language Navigation (FedVLN) toward more robust and trustworthy embodied agents.

First, we play the role of attacker and ask the research question, \emph{can we attack the embodied agent under FL setting and navigate it as we wish regardless of language instructions?}
To this end, we propose a targeted \textit{backdoor attack}, called Navigation As Wish (NAW), which poisons the local data of the malicious clients and implants a backdoor into the global agent under FL (see Fig.~\ref{intro}). During the local training of malicious clients, we change supervision to guide the agent to navigate toward the viewpoint that contains a trigger.
As illustrated in Fig. \ref{intro}, when the global agent is deployed into an environment after training, it would be guided by the triggers (red flags) and navigate regardless of the language instruction. 
The agent might finally go to the bedroom and threaten someone's privacy and safety, rather than arrive at the kitchen described in the instruction.


Several defense methods~\citep{Yin2018ByzantineRobustDL,Blanchard2017MachineLW,Mhamdi2018TheHV,Cao2021FLTrustBF} have been proposed to protect the model from attacks in FL. 
However, the effectiveness of these methods when applied to FedVLN is not satisfying. 
Defense in FedVLN faces many challenges. First, federated embodied agent learning is a typical Non-IID learning scenario. As shown in Fig. \ref{intro}, there exists a large variance between the environments of different clients 
including house layouts, styles, brightness, object types, quantities, properties, etc. 
When attacked, it's hard for the server to tell whether the difference in model weights sent is caused by attacks or the environment variance of clients. Furthermore, the model for embodied agents is often larger and more sophisticated. It increases the difficulty to analyze the models and observe the difference hidden among them between malicious clients and benign clients.

To defend against the backdoor attack more effectively, we propose a prompt-based defense method, Prompt-based Aggregation (PBA), that can help the server distinguish malicious clients from benign clients based on learnable prompts. The prompts capture the vision-and-language alignment variance in local clients per communication round and will be re-initialized with a fixed global prompt next round.
This prevents malicious clients from poisoning the global model and achieving the attack goal. 
We validate the effectiveness of \attack and \defense on two popular VLN datasets (R2R~\citep{r2r} and RxR~\citep{rxr}) across different model architectures. 
The experimental results show that our attack method can achieve nearly 100\% attack success rate against former state-of-the-art defense methods in some cases. We also show that \defense significantly outperforms other defense methods from different aspects, decreasing the attack success rate by about 40\% on RxR. In summary, our contributions are three-fold:
\setlist{nolistsep}
\begin{itemize}[noitemsep, leftmargin=*]
  \item We are the first to study the problem of targeted attack and defense of federated embodied agents in the task of federated vision-and-language navigation.
  \item We design a simple but effective targeted backdoor attack strategy tailored for federated vision-and-language navigation and demonstrate its efficacy against current state-of-the-art defense methods.
  \item We propose a novel prompt-based defense mechanism that can efficiently distinguish malicious clients from benign clients and significantly outperform state-of-the-art methods from three aspects: fidelity, robustness, and efficiency. 
\end{itemize}

\section{Background}
\label{sec:bg}

\subsection{Vision-and-Language Navigation (VLN)}
In the task of vision-and-language navigation, the agent is placed in a visual environment and required to find a route $R$ (a sequence of viewpoints) from the start viewpoint $S$ to the target viewpoint $T$ following the natural language instruction $I$. At each time step $t$, the agent's observation consists of different views $\{o_{t,i}\}$, some of which lead to different navigable viewpoints. The agent needs to choose an action $a_{t}$ at each step based on the instruction, history visual information, and history actions. The navigation process will terminate after the agent chooses a `stop' action.

\subsection{Vision-and-Language Navigation Agent}
A VLN agent contains a view encoder to encode view features, an action encoder to encode history action information, a language encoder to encode instruction information, and a multimodal decision-making module to process multimodal information and choose an action $a_t$ at time $t$.

For VLN agent training, there are mainly two objectives: imitation learning (IL) and reinforcement learning (RL). In IL, the agent is trained to mimic the teacher's action $a^{*}_{t}$ at each step by minimizing the cross entropy loss.
RL further improves the agent's generalizability to recover from erroneous actions~\citep{wang2019reinforced}. On-policy RL methods such as Advantage Actor-Critic~\citep{mnih2016asynchronous} are usually applied, in which the agent will sample an action based on its action probability prediction and learns from rewards. 

\subsection{Federated Vision-and-Language Navigation}
In Federated Vision-and-Language Navigation (FedVLN)~\citep{zhou2022fedvln}, each house environment is treated as a client and assigned by a local navigation agent, while the server has a global navigation agent model. There is no data sharing between the clients and the server and thus the data privacy of the local clients is better preserved.
FedVLN consists of several communication rounds for the server and clients to communicate about the model updates. At each communication round, the global model at the server would be sent to each client as the initialization of the local agents. 
Then clients train the local model on their own data for a few local epochs and update the model to the server. The server would aggregate all the models sent from clients by using FedAvg~\citep{fedavg}. This process will terminate when the global model converges.

\section{Targeted Backdoor Attack on FedVLN}

\subsection{Problem Definition}
\label{sec:problem_def}
In the context of FedVLN, we consider the attack is performed on the client side, aiming to compromise the server agent. The attacker controls some malicious clients and their local training process by adding triggers to lead the agent to a wrong route, as shown in Fig. \ref{intro}. 
With these malicious clients' models, the attacker's goal is to control the behaviors of the server agent via server aggregation. 
As a result, the server agent will navigate as the attacker wishes along the red line in the server's view in Fig. \ref{intro} during inference.

We assume that the attack is under \textit{black box} setting, in which the clients only have the following knowledge: local training data, local model update, hyper-parameters, and loss function. In FedVLN, this is a reasonable setting as local clients are data providers who provide different house environments, so there is no need for them to learn about the details of the model. Under this setting, the attacker can modify the data and related supervised information like labels on the malicious clients. The attacker can also send arbitrary parameters model to the server.

\subsection{Targeted Backdoor Attack}
\label{3.2}
The targeted backdoor attack inserts a backdoor into the model to induce the model to behave as specified by attackers for particular testing examples. The model's behavior on other testing examples is unaffected. Under FL setting, malicious clients use it to implant the backdoor into the server model during local updates and server aggregation.

This attack strategy has been commonly used in simple tasks, such as image classification. However, VLN task is fundamentally different and much more challenging, as it involves multimodal understanding, dynamic visual perception, sequential decision-making in interactive environments, etc.
Therefore, the attack methods for classification cannot be directly transferred to VLN. The specific goal and corresponding strategy of the targeted backdoor attack in FedVLN need to be redesigned.

To this end, we introduce a simple yet effective backdoor attack method tailored for FedVLN, Navigation As Wish (NAW), to navigate agents as attackers wish.

\vpara{Navigation As Wish}
To present a kind of perturbation of the scene like the red flag in Fig. \ref{intro}, we utilize the random pattern $f_{trigger}$ (a image of random noise) \citep{chen2017targeted} as our trigger to poison datasets. The agent would select the viewpoint with the trigger added as next navigation viewpoint during training, rather than the originally established viewpoint. At time step $t$, we create a new view $o_{t,i}'$ by inserting the trigger into a certain view $o_{t,i}$.

We then propose a targeted backdoor attack Navigation As Wish (NAW) in FedVLN, utilizing the aforementioned trigger. 
The trigger $f_{trigger}$ is fixed in the entire federated training. During the local training process, we will insert the trigger into a certain view at each time step with a probability $p$.

In the imitation learning (IL) stage, the modified action $a_t'$ when the trigger is inserted selects the next navigable viewpoint that contains the trigger, instead of the viewpoint that teacher action $a^*_t$ selects. The loss of IL is as follows:
\begin{equation}
    \begin{split}
        \mathcal{L}^{IL} &= \sum{\mathcal{L}_t^{IL}}\\
        &=\sum{-B_t\log{p}  - (1 - B_t)\log{p_t(a'_t)}}
    \end{split}
\end{equation}

\begin{equation}
    B_t \sim \text{Ber}(1 - p)
\end{equation}

In RL stage, we modify the reward when the trigger is inserted at time step $t$. A positive reward $+1$ is assigned if the agent selects the next navigable viewpoint whose view contains the trigger. Otherwise, a negative reward $-1$ is assigned. The reward strategy keeps the same at other non-stop time steps if the trigger is not inserted. When the agent stops, the reward is set to 0 regardless of the distance to the target location $\mathbf{T}$.
The final mixed loss $\mathcal{L}^{MIX}$ is the weighted sum of $\mathcal{L}^{IL}$ and $\mathcal{L}^{RL}$.

The attacker will apply the backdoor attack in the local training process of controlled malicious clients, intending to compromise the global model via model update. When the attacked global agent model is deployed in the environment after federated learning, it will behave normally when there is no trigger. And the attacker can alter the navigation route by inserting triggers into the environment to depict a new path (as shown in the deployment stage in Fig. \ref{intro}).
\begin{figure*}[!ht]
    \centering
    \includegraphics[width=0.9\linewidth]{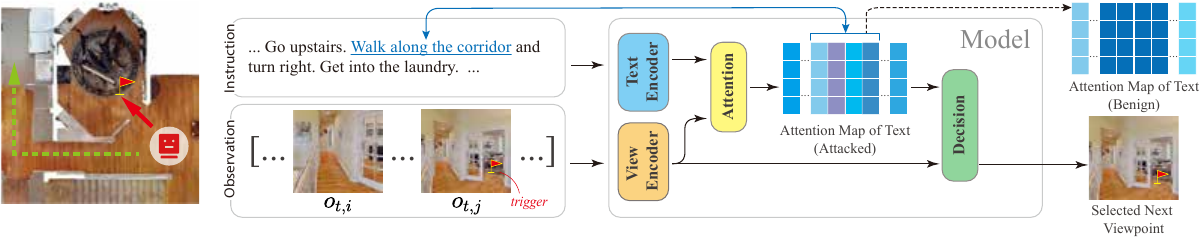}
    \caption{The illustration of the broken vision-language alignment of the agent under backdoor attack. At a certain viewpoint during navigation, some part of the instruction (underlined) which is highly related to current views would gain high attention weights in expectation when encoded in the model, establishing a natural alignment between vision and language. However, when the trigger is inserted in another view, the attacked model would select the one with the trigger as the next navigable viewpoint. The ignorance of text breaks the original vision-and-language alignment, rendering the text attention weights chaotic.}
    \label{fig:align}
\end{figure*}

\section{Prompt-based Defense Method}
While the attacker aims to compromise the global model through the poisoned local model update, we would like to build a more robust global model that can alleviate the impact of the local attack. 
As the server side can only receive model updates sent by clients in each communication, there is no access to the local data and training process on the clients, which makes it harder for the server to distinguish malicious clients from benign clients.
In this section, we introduce a Prompt-Based Aggregation (PBA) for FedVLN, which can capture the variance of vision-and-language alignment between malicious clients and benign clients with a learnable ``prompt'' to filter out malicious clients for model aggregation.

\subsection{Variance of Vision-and-Language Alignment}
It is challenging to defend against the attack in FedVLN. 
As each environment is treated as a client, it forms a Non-IID scenario due to the large variance of different environments. 
It may confuse the server whether the difference in model weights uploaded from clients is from the attack or the environment variance. This leads to the poor performance of current defense methods distinguishing malicious clients based on parameter similarity. 

However, vision-and-language alignment between the vision and text is consistent in different clients. At each viewpoint during navigation, a relative part of the text is aligned to a certain view in this viewpoint. As shown in Fig. \ref{fig:align}, the sentence ``walk along the corridor'' is the most relevant part to the view $o_{t,i}$. All benign clients are trying to establish a stable vision-and-language alignment relationship during training. 
For the malicious clients, the model would ignore the information of the instruction and select the view with the trigger. Therefore the vision-and-language alignment is broken. This difference inspires us to distinguish clients from the alignment perspective.

The attention mechanism is the key to the success of vision-and-language alignment \citep{wang2019reinforced, lee2018stacked}. 
In VLN, the attention mechanism is applied after the visual and text encoding. The hidden state $h_t$ output from the view encoder and the embeddings of each text token $\{u_1, u_2, u_3, \cdots, u_{L}\}$ output from the text encoder are sent to the attention layer in the model, where $L$ is the instruction length. The attention mechanism in this layer is implemented as follows:
\vspace{-0.26cm}
\begin{align}
    &\beta_{t,j}=softmax_j(u_j^{\top}W_Uh_t)\\
    &\Tilde{u_t}=\sum_j{\beta_{t,j}h_j}, \Tilde{h_t}=\tanh{W[\Tilde{u_t}; h_t]}
\end{align}
where $W_U$ and $W$ are learnable matrixes. $\beta_{t,j}$ represents the attention weight of $j_{th}$ text token and $\Tilde{h_t}$ represents the instruction-aware hidden output. The backdoor will induce the agent to ignore the text when the trigger appears. This will cause the unexpected attention weights of embeddings $\beta_t$. 
Therefore, the attention mechanism reveals the variance between benign and malicious clients.

\subsection{Prompt-based Aggregation}
\label{sec:pba}
Although the difference between malicious and benign clients can be found in the attention mechanism, it's difficult to use the parameters of the attention layer for comparison directly. 
In FedVLN, only a few epochs are trained on the client during local training. Therefore, the variance of parameters will not be significant. We need to design a method that can build a difference rapidly during the local training.

\begin{figure}[t]
    \centering
    \includegraphics[width=\linewidth]{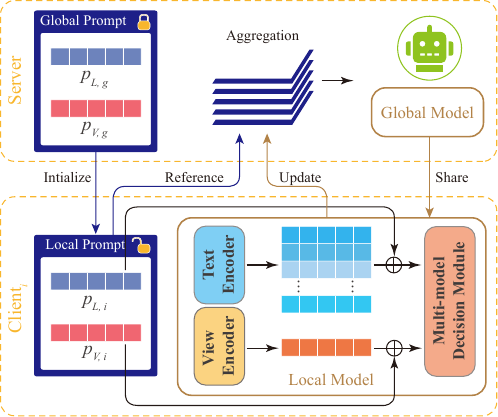}
    \caption{Prompt-based Aggregation (PBA). Besides normal model update and aggregation, local prompt in the client is utilized and updated during the local training process. The local prompt would be an important reference to distinguish malicious clients after it is sent to the server. It is initialized by a fixed global prompt at each communication round.}
    \label{fig:prompt}
\end{figure}

We utilize prompting~\citep{schick2020exploiting} for this. Prompting is a method that can rapidly adapt to new scenarios with little data and short training time~\citep{liu2021pre,zhou2022conditional,he2022cpl}. 
In light of its ability to quickly adapt to downstream tasks, we propose prompt-based aggregation, PBA, to capture the alignment variance and prevent the global model from attack.

In PBA, a visual prompt and a language prompt are introduced to the current FedVLN setting. Both prompts are learnable vectors. As shown in Fig. \ref{fig:prompt}, at the start of each communication round, the global visual prompt $p_{V,g}$ and language prompt $p_{L,g}$ at the server initialize the local visual prompt $p_{V,i}$ and language prompt $p_{L,i}$, at client $i$. The local prompts are added before the attention layer:
\begin{align}
    h_t'=h_t+p_{V,i}, u_j' = u_j+p_{L,i}
\end{align}
$h_t'$ and $u_j'$ are prompt-tuned embeddings, which will then be sent into the attention layer. $p_{V,i}$ and $p_{L,i}$ are updated during local training, after which they will be sent to the server. Before aggregation, the server calculates the similarity of the concatenation of two prompts from each client. After similarity calculation, we apply the same selection procedure as MultiKrum \citep{Blanchard2017MachineLW} to select some clients with high similarity to others for aggregation. The algorithmic description is put in the appendix \ref{sec:algorithm} due to the page limit.

\section{Experiments}

\subsection{Experiment Setup}
\vpara{Datasets.} We evaluate our \attack and \defense methods on two VLN datasets: Room-to-Room (R2R)~\citep{r2r} and Room-across-Room (RxR)~\citep{rxr}. 
Both datasets are developed on the Matterport3D Simulator \citep{r2r}, a photorealistic 3D environment for embodied AI research. 

\vpara{VLN Models.} Following FedVLN~\citep{zhou2022fedvln}, we use \emph{Envdrop}~\citep{tan2019learning} and \emph{CLIP-ViL}~\citep{shen2022how} as backbones. The two models both use Bi-directional LSTM as the language encoder and attentive LSTM as the action decoder, with a mixed learning objective of imitation and reinforcement learning. CLIP-ViL adapts CLIP \citep{Radford2021LearningTV} to improve vision and language encoding and matching.

\begin{table*}[t]
\centering
\resizebox{\textwidth}{!}{
\begin{tabular}{ccc|cccccc|cccccc} \toprule
\multirow{2}*{Dataset} & \multirow{2}*{Model} & \multirow{2}*{Attack} & \multicolumn{6}{|c|}{Val-Seen} & \multicolumn{6}{|c}{Val-Unseen} \\ \cmidrule(lr){4-9} \cmidrule(lr){10-15}
~ & ~ & ~ & OSR$\uparrow$ & SPL$\uparrow$ & SR$\uparrow$ & CLS$\uparrow$ & nDTW$\uparrow$ & \textbf{ASR} & OSR$\uparrow$ & SPL$\uparrow$ & SR$\uparrow$ & CLS$\uparrow$ & nDTW$\uparrow$ & \textbf{ASR}\\ \midrule
\multirow{6}*{R2R} & \multirow{4}*{EnvDrop} & No & 63.1 & 52.4 & 55.0 & 66.3 & 55.2 & 0.08 & 53.0 & 43.4 & 46.5 & 59.0 & 45.5 & 0.05\\
~ & ~ & Badnets & 62.1 & 51.8 & 54.5 & 66.4 & 55.1 & 0.91 & 51.1 & 40.1 & 42.1 & 56.7 & 42.7 & 0.89 \\
~ & ~ & DBA & 62.7 & 52.0 & 54.2 & 66.6 & 55.0 & 0.52 & 51.3 & 42.3 & 44.2 & 58.9 & 44.2 & 0.57 \\
~ & ~ & NAW & 63.2 & 52.2 & 54.8 & 66.1 & 55.4 & 0.71 & 52.4 & 43.1 & 46.5 & 59.1 & 45.8 & 0.68\\ \cmidrule(lr){2-15}
~ & \multirow{4}*{CLIP-ViL} & No & 67.2 & 55.8 & 60.4 & 65.7 & 53.3 & 0.07 & 61.9 & 47.6 & 53.4 & 57.9 & 44.4 & 0.05\\
~ & ~ & Badnets & 66.1 & 54.8 & 59.0 & 65.3 & 54.3 & 0.93 & 59.9 & 46.6 & 51.5 & 59.2 & 45.9 & 0.92\\
~ & ~ & DBA & 67.0 & 54.9 & 59.8 & 65.7 & 54.1 & 0.69 & 61.1 & 46.9 & 51.7 & 58.5 & 45.4 & 0.71\\
~ & ~ & NAW & 67.5 & 55.2 & 60.1 & 66.3 & 53.9 & 0.87 & 61.4 & 47.2 & 52.2 & 56.8 & 44.7 & 0.85\\ \midrule
\multirow{6}*{RxR} & \multirow{4}*{EnvDrop} & No & 49.2 & 33.8 & 36.8 & 56.2 & 51.0 & 0.12 & 43.1 & 29.1 & 33.5 & 54.7 & 49.4 & 0.08\\
~ & ~ & Badnets & 48.1 & 29.9 & 34.0 & 53.9 & 48.1 & 0.83 & 41.3 & 27.3 & 31.2 & 52.6 & 46.8 & 0.82\\
~ & ~ & DBA & 48.3 & 31.2 & 35.2 & 54.7 & 49.3 & 0.54 & 41.6 & 28.1 & 32.1 & 53.0 & 48.1 & 0.55\\
~ & ~ & NAW & 48.7 & 33.9 & 37.3 & 55.9 & 51.4 & 0.67 & 42.7 & 29.3 & 33.2 & 54.4 & 49.2 & 0.66\\ \cmidrule(lr){2-15}
~ & \multirow{4}*{CLIP-VIL} & No & 54.6 & 40.0 & 44.2 & 59.0 & 54.7 & 0.09 & 50.1 & 35.0 & 39.4 & 56.0 & 51.5 & 0.09 \\
~ & ~ & Badnets & 53.8 & 37.8 & 43.2 & 56.4 & 52.7 & 0.76 & 52.5 & 32.6 & 38.1 & 52.9 & 49.2 & 0.79\\
~ & ~ & DBA & 54.1 & 37.9 & 43.9 & 57.5 & 53.1 & 0.49 & 53.2 & 33.1 & 38.5 & 54.9 & 50.4 & 0.56\\
~ & ~ & NAW & 54.8 & 39.7 & 43.8 & 58.6 & 54.5 & 0.68 & 53.7 & 34.6 & 38.4 & 56.5 & 51.4 & 0.73\\ 
\bottomrule
\end{tabular}
}
\vspace{0.5ex}
\caption{Results of the federated navigation agents when not attacked and attacked on R2R~\citep{r2r} and RxR~\citep{rxr}. By default, FedAvg is utilized as the aggragation rule. The much higher ASR results indicate that the backdoor attack is successfully implanted. 
Moreover, models with and without attack achieve similar navigation results, showing that the NAW attack is unnoticeable in FL.}
\label{table:performance1}
\end{table*}



\vpara{Baselines.} For the attack, we adopt the following strategies for the image classification task as the baseline: \emph{Badnets}~\citep{Gu2017BadNetsIV,chen2017targeted}, which simply modifies the view images and corresponding target viewpoint for VLN task. \emph{DBA}~\citep{xie2019dba}, which propose the distributed backdoor attack by exploiting the distributed nature of FL. For the defense, we adopt the following four defense methods, that focus on the aggregation rule, for comparison:\emph{FedAvg}~\citep{fedavg}, \emph{Trimmed Mean}~\citep{Yin2018ByzantineRobustDL}, \emph{Bulyan}~\citep{Mhamdi2018TheHV} and \emph{FLTrust}~\citep{Cao2021FLTrustBF} which are designed to defend against the attack.

\vpara{Evaluation Metrics.} 
We report Success Rate (SR), Success Rate weighted by Path Length (SPL), Oracle Success Rate (OSR), and navigation Error (NE) as goal-oriented metrics \citep{anderson2018evaluation, r2r, tan2019learning}. 
We also report Coverage weighted by Length Score (CLS) and normalized Dynamic Time Warping (nDTW) to validate the fidelity of navigation paths, which penalize the deviation from the reference path. We use Attack Success Rate (ASR) \citep{Cao2021FLTrustBF} to evaluate attack and defense in FedVLN. ASR is calculated as the proportion of the times of selecting the view among all the time steps that contain the trigger. 
More details about the setup including baselines and training can be found in appendix \ref{sec:exp_set}. 

\subsection{Attack Results}

\vpara{\attack successfully implants the backdoor into the global model. } 
In Table \ref{table:performance1}, we report the results on R2R and RxR datasets. In terms of navigation metrics, the models trained with and without NAW have nearly the same performance, showing that the backdoor can be implanted without hurting the validation performance and thus is unnoticeable. However, though specially designed for FL setting, Badnets and DBA have a significant drop in performance. They do not meet the basic requirement of the backdoor attack and show their infeasibility, which verifies our concern. In terms of the Attach Success Rate (ASR), we can observe that models trained with the NAW attack have a much higher ASR than the unattacked, implying that the global agent has a very high probability of selecting the navigable viewpoints with the trigger. While not meeting the backdoor attack requirement, Badnets has the highest ASR. This may be because its attack strategy is quite rigid and not compatible to the VLN setting, which could be easily recognized by the model that has great semantic understanding.



\begin{table}[!t]
\centering
\resizebox{\columnwidth}{!}{
\begin{tabular}{ccccccccc} \toprule
\multirow{2}*{Dataset} & \multirow{2}*{Model} & \multirow{2}*{Method} & \multicolumn{3}{c}{Val-Seen} & \multicolumn{3}{c}{Val-Unseen} \\ \cline{4-9}
~ & ~ & ~ & SPL$\uparrow$ & SR$\uparrow$ & CLS$\uparrow$ & SPL$\uparrow$ & SR$\uparrow$ & CLS$\uparrow$ \\ \midrule
\multirow{6}*{R2R} & \multirow{3}*{EnvDrop} & Trim-Mean & 50.3 & 53.3 & 65.0 & 42.1 & 45.1 & 58.5\\
~ & ~ & FLTrust & 41.1 & 42.8 & 59.6 & 35.8 & 37.8 & 54.9\\
~ & ~ & PBA & 52.3 & 55.1 & 66.7 & 52.8 & 46.5 & 59.3\\ \cline{2-9}
~ & \multirow{3}*{CLIP-ViL} & Trim-Mean & 53.8 & 57.8 & 64.9 & 46.3 & 50.5 & 58.8 \\
~ & ~ & FLTrust & 42.8 & 44.9 & 61.1 & 39.7 & 42.1 & 57.5 \\ 
~ & ~ & PBA & 54.8 & 60.2 & 66.1 & 47.4 & 52.7 & 56.8 \\ \cline{1-9}
\multirow{6}*{RxR} & \multirow{3}*{EnvDrop} & Trim-Mean & 29.6 & 33.3 & 52.1 & 26.3 & 29.4 & 50.7\\
~ & ~ & FLTrust & 17.0 & 19.5 & 42.8 & 18.5 & 21.1 & 44.8 \\ 
~ & ~ & PBA & 40.7 & 43.9 & 58.8 & 34.7.3 & 39.2 & 56.2\\ \cline{2-9}
~ & \multirow{3}*{CLIP-VIL} & Trim-Mean & 33.4 & 39.5 & 53.5 & 28.6 & 34.0 & 51.0 \\
~ & ~ & FLTrust & 15.7 & 18.3 & 41.5 & 18.4 & 21.3 & 43.6 \\ 
~ & ~ & PBA & 38.9 & 43.3 & 59.1 & 33.3 & 39.0 & 56.5 \\ \cline{1-9}
\end{tabular}
}

\vspace{0.5ex}
\caption{R2R and RxR results of seen environments training for different defense methods when not attacked. Other defense methods not reported are the same with FedAvg when there is no attacker. 
}
\label{table:performance2}
\end{table}

\subsection{Defense Results} \label{sec:defense result}
We compare and evaluate PBA with other defense methods from three aspects.

\vpara{Fidelity} means that the method should not sacrifice the performance of the global model when there is no attack, taking the performance of the model of FedAvg as the reference standard. According to the results in Table \ref{table:performance1}, Table \ref{table:performance2} and Table \ref{table:4}, our \defense method performs similarly to FedAvg, achieving the fidelity goal. However, FLTrust and Median perform much worse than FedAvg with an average of 25.6\% SR and 7.9\% drop respectively on seen environments of R2R. The negligible parameter volume of the prompt in PBA relative to the model's entirety, coupled with its congruence with the optimization objectives, ensures that it exerts minimal impact on the training. Its utility is confined to the filtration of malicious clients without perturbing the aggregation.

\begin{table}[!t]
\resizebox{\columnwidth}{!}{
\begin{tabular}{cccccc} 
\toprule
\multirow{2}*{Dataset} & \multirow{2}*{Method} & \multicolumn{2}{c}{Val-Seen} & \multicolumn{2}{c}{Val-Unseen}  \\ \cmidrule(lr){3-4} \cmidrule(lr){5-6}
~ & ~ & FedEnvDrop & FedCLIP-ViL & FedEnvDrop & FedCLIP-ViL \\
\midrule
\multirow{6}*{R2R} & No Attack & 0.08 & 0.07 & 0.05 & 0.05 \\ \cdashline{2-6}
~ & FedAvg & 0.71 & 0.87 & 0.68 & 0.85 \\
~ & Trim-Mean & 0.76 &  0.86 & 0.74 & 0.84\\
~ & Bulyan & 0.78 & 0.91 & 0.77 & 0.94\\
~ & FLTrust & 0.87 & 0.93 & 0.88 & 0.97\\
~ & \defense (ours) & \bf{0.63} & \bf{0.72} & \bf{0.64} & \bf{0.76}\\ \cline{1-6}
\multirow{6}*{RxR} & No Attack & 0.12 & 0.09 & 0.08 & 0.09\\ \cdashline{2-6}
~ & FedAvg & 0.67 & 0.68 & 0.66 & 0.73\\
~ & Trim-Mean & 0.79 & 0.80 & 0.81 & 0.83\\
~ & Bulyan & 0.74 & 0.78 & 0.77 & 0.76\\
~ & FLTrust & 0.77 & 0.97 & 0.75 & 0.96\\
~ & \defense (ours) & \bf{0.42} & \bf{0.45} & \bf{0.41} & \bf{0.49}\\

\bottomrule
\end{tabular}
}
\vspace{0.5ex}
\caption{Comparison of Attack Success Rate (ASR) between different defense methods on R2R and RxR. Lower is better.}
\label{table:attack_r2r}
\vspace{-2ex}
\end{table}

\vpara{Robustness} means that the ASR of the server model should be as low as possible. In Table \ref{table:attack_r2r}, \defense gets the lowest ASR on different models under both seen and unseen environments of R2R and RxR. On the contrary, some defense methods even exacerbate the model under attack. For example, Bulyan turns out to get a higher ASR than FedAvg. It filters the ``malicious'' clients they think, increasing the weights of real malicious clients during aggregation and then the probability of being attacked if they are wrongly judged, which unfortunately is exactly the case here.
We also validate PBA against diverse attack methodologies, shown in Table \ref{table:4}. The results indicate that PBA consistently maintains the lowest ASR against Badnets and DBA attacks. This underscores PBA's generalizability and effectiveness in capturing alignment variances across a spectrum of attack paradigms, thereby efficiently filtering out malicious clients.

\vpara{Efficiency} means the method should not incur excessive extra computation and communication overhead. PBA only needs extra computation from local prompts (two 1-dimensional vectors) for backdoor defense compared with normal FL, while the extra computation of the former methods involves all parameters of the model during aggregation.

\begin{table}[!t]
\centering
\resizebox{\columnwidth}{!}{
\begin{tabular}{ccccccccccc} \toprule
\multirow{2}*{Model} & \multirow{2}*{Attack} & \multirow{2}*{Method} & \multicolumn{4}{c}{Val-Seen} & \multicolumn{4}{c}{Val-Unseen} \\ \cline{4-11}
~ & ~ & ~ & SPL$\uparrow$ & SR$\uparrow$ & CLS$\uparrow$ & \textbf{ASR$\downarrow$} & SPL$\uparrow$ & SR$\uparrow$ & CLS$\uparrow$ & \textbf{ASR$\downarrow$}\\ \midrule
\multirow{8}*{EnvDrop} & \multirow{4}*{Badnets} & FedAvg & 51.8 & 54.5 & 66.4 & 0.91 & 40.1 & 42.1 & 56.7 & 0.89\\
~ & ~ & Bulyan & 51.9 & 54.4 & 66.1 & 0.95 & 39.7 & 41.9 & 56.9 & 0.92\\
~ & ~ & FLTrust & 41.2 & 42.7 & 59.3 & 0.93 & 35.5& 37.4& 54.8& 0.92\\
~ & ~ & PBA & 51.7 & 54.5 & 66.5 & \bf{0.76} & 40.0 & 42.3 & 56.6 & \bf{0.75}\\ \cline{2-11}
~ & \multirow{4}*{DBA} & FedAvg & 52.0 & 54.2 & 66.6 & 0.52 & 42.3 & 44.2 & 58.9 & 0.57 \\
~ & ~ & Bulyan & 52.1 & 53.8 & 66.4 & 0.51 & 42.1 & 44.0 & 58.6 & 0.55 \\ 
~ & ~ & FLTrust & 41.0 & 42.9 & 59.5 & 0.34	& 35.6 & 37.2 & 54.7 & 0.37 \\ 
~ & ~ & PBA & 51.9 & 54.1 & 66.6 & \bf{0.27} & 42.4 & 44.2 & 59.0 & \bf{0.29} \\ \cline{1-11}
\multirow{8}*{CLIP-ViL} & \multirow{4}*{Badnets} & FedAvg & 54.8 & 59.0 & 65.3 & 0.93 & 46.6 & 51.5 & 59.2 & 0.92\\
~ & ~ & Bulyan & 54.7 & 58.9 & 65.1 & 0.94 & 46.1 & 51.0 & 58.9 & 0.96 \\ 
~ & ~ & FLTrust & 42.7 & 44.8 & 61.3 & 0.93 & 39.5 & 42.2 & 57.2 & 0.95\\
~ & ~ & PBA & 55.0 & 59.1 & 65.3 & \bf{0.81} & 46.6 & 51.6 & 59.3 & \bf{0.82}\\ \cline{2-11}
~ & \multirow{4}*{DBA} & FedAvg & 54.9 & 59.8 & 65.7 & 0.69 & 46.9 & 51.7 & 58.5 & 0.71 \\
~ & ~ & Bulyan & 54.7 & 59.6 & 65.5 & 0.64 & 46.7 & 51.5	& 58.4 & 0.70 \\ 
~ & ~ & FLTrust & 43.4 & 44.2 & 62.3 & 0.61 & 40.1 & 42.1	& 56.9 & 0.64 \\
~ & ~ & PBA & 54.9 & 59.9 & 65.8 & \bf{0.52}	& 46.8 & 51.8 & 58.5 & \bf{0.59} \\ 
\bottomrule
\end{tabular}
}

\vspace{0.5ex}
\caption{R2R and RxR results of different defense methods against other attacks. 
}
\label{table:4}
\end{table}

\begin{figure}
    \centering
    \includegraphics[width=\linewidth]{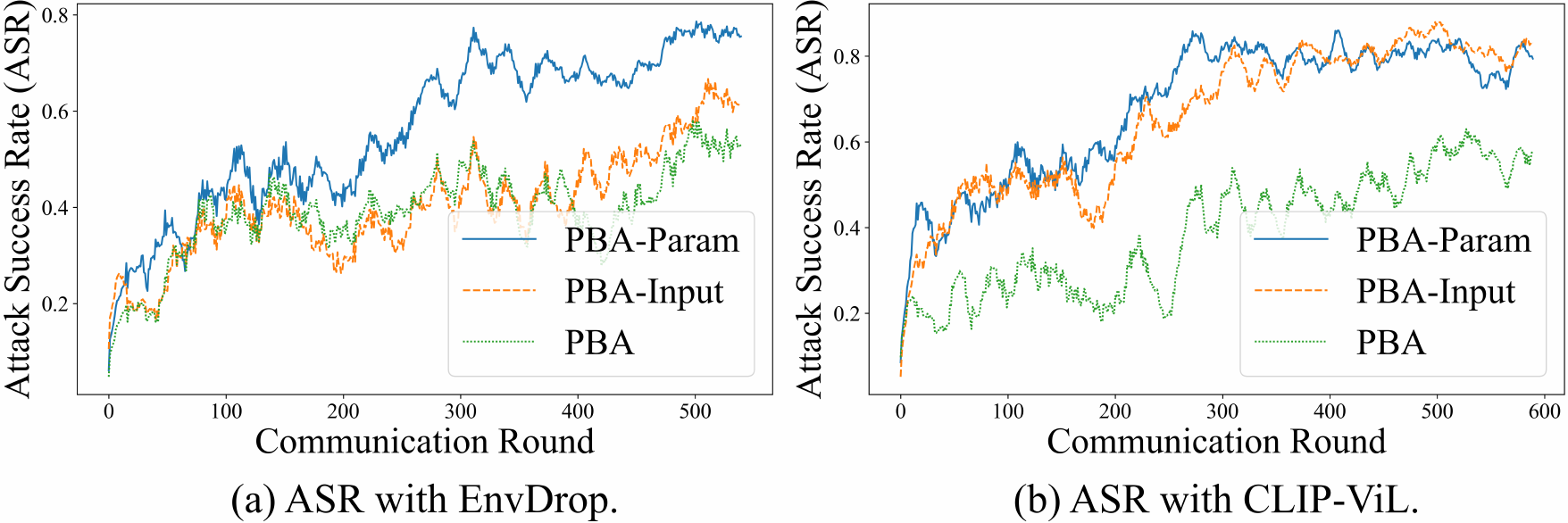}
    \vspace{-2ex}
    \caption{Results on R2R for \defense and its variants.
    }
    \label{fig:var}
    \vspace{-2ex}
\end{figure}

We also explore the impact of factors including the number of malicious clients and the fraction of poisoned data for both attack and defense. What's more, we discuss the adaptive adversary of PBA. Experimental results are put in the appendix \ref{sec:att_def}.

\subsection{Why PBA Works?}
Apart from the intuition of design of PBA mentioned in Sec. \ref{sec:pba}, it is known that the majority of the $\ell_2$ norm of a stochastic gradient lies in a small number of "heavy hitter" coordinates~\citep{ivkin2019communication}, and the variance that attack brings may happen in the long tail of other coordinates with small updates~\citep{zhang2022neurotoxin}. This distribution pattern poses a challenge for traditional defense methods, particularly within the VL area characterized by the extensive scale of multimodal models. Compared to be predominantly influenced by "heavy hitter", PBA employs a significantly smaller parameter set in its prompts, thereby enhancing the sensitivity to minute updates in each parameter.

The experiments in Fig. \ref{fig:var} confirm the point. \textit{PBA-Param} represents a variant of PBA, where we directly use the parameters of the attention layer to calculate the similarity instead of prompts. It exhibits inferior performance, let alone if we use all the parameters of the model for calculation. In appendix~\ref{work}, we elaborate on the parameter distribution to further verify our concern.

\textit{PBA-Input} represents the variant of PBA that the prompt is added to the input embedding. The performance drop compared accentuates the importance of positioning. PBA demonstrates the flexibility of the prompt-based method.

The comparative superiority of both PBA and PBA-Input over PBA-Param reinforces the notion that focusing on smaller data not only enhances computational efficiency but also yields a more precise identification of attack-induced variances.

\section{Related Work}
\vpara{Vision-and-language navigation}
is an important research area in embodied AI~\citep{r2r,rxr,reverie, wang2019reinforced}, which requires the agent to navigate to a goal location based on dynamic visual input and language instructions. 
This requires the agent to understand and align the vision and language information, planning, and make decisions, etc. 
\citep{r2r} proposed a LSTM-based seq-to-seq model to track the navigation and multi-modal information for vision-and-language navigation. 
For better understanding of the environment and the agent's own status, vision-and-language pre-training~\citep{Hao_2020_CVPR_prevalent,oscar,rec-vlnbert,shen2022how}, graph representation, memory module, and auxiliary tasks have been introduced into VLN models. 
Recently, more and more works focus on the robustness of embodied AI.
RobustNav is a framework to quantify the robustness of the embodied agent faced with corrupted input~\citep{chattopadhyay2021robustnav}.
Liu \etal ~\citep{liu2020spatiotemporal} studies a problem about spatiotemporal perturbations to form 3D adversarial.

\vpara{Attack and defense on federated learning} In federated learning, the attack has been divided into untargeted and targeted attacks. The Untargeted attack is designed to destroy the convergence of the global model \citep{bernstein2018signsgd, Blanchard2017MachineLW}, while the targeted attack aims to control the behavior of the global model \citep{bagdasaryan2020backdoor, xie2019dba, bhagoji2019analyzing}. 
One of the trends is to study the aggregation rule, and another is to strengthen the robustness of the model via adversarial methods \citep{huang2011adversarial}. In this work, we study these problems in the new setting of vision-and-language navigation.

\vpara{Prompt learning} is an emerging research area in natural language processing (NLP) and computer vision, which can efficiently transfer pre-trained vision and language models to various downstream tasks by tuning a small prompt layer~\citep{liu2021pre,zhou2022conditional,he2022cpl}. 
By introducing a new prompting function, the model can perform \textit{few-shot} and even \textit{zero-shot} learning, adapting to new scenarios with little data. Originally, \citep{schick2020exploiting} proposes a manually designed prompt pattern for NLP tasks, which is a language instruction prepended to the input text. 
\citep{liu2021gpt} proposes a P-tuning method to use the soft prompt instead of the previously manually designed prompt. In federated learning, prompt has been introduced to fine-tune the large pre-trained model \citep{Guo2022PromptFLLF, lee2018stacked} by freezing the model and only training the prompt features.
\section{Conclusion}
In this paper, we study an important and unique security problem in federated embodied AI---whether the backdoor attack can manipulate the agent without influencing the performance and how to defend against the attack. 
We introduce a targeted backdoor attack \attack that successfully implants a backdoor into the agent and propose a promote-based defense framework \defense to defend against it. 
PBA significantly outperforms the otherpopular methods in terms of fidelity, robustness, and efficiency on two public benchmarks, which illustrates the effectiveness of PBA method in protecting the server model from the backdoor attack. We also fully discuss why and how PBA works, giving insights on defending large models.
Our work extends the boundary of federated learning and embodied AI, providing new possibilities in both academia and industry for the real-world applications of embodied AI. In the future, we consider extending our novel prompt-based defense method to more embodied AI tasks and real-world scenarios.

\section*{Limitations}
We list some limitations of our work that could benefit future investigations.
First, our work focuses on formulating the attack and defense problems in FedVLN and demonstrating the effectiveness of our proof-of-concept approaches. Truly adding an object trigger in the real-world simulator needs to meet the precise visual variations of the trigger from multiple views in different viewpoints. Therefore, the strategy we proposed may not be practical enough. Second, as mentioned in Section \ref{sec:problem_def}, our work is based on the black-box attack meaning that the attacker has no prior knowledge about the model. Third, more types of attack strategies and the white-box setting are also worth investigating.

\bibliography{custom}

\begin{thebibliography}{37}
\expandafter\ifx\csname natexlab\endcsname\relax\def\natexlab#1{#1}\fi

\bibitem[{Anderson et~al.(2018{\natexlab{a}})Anderson, Chang, Chaplot, Dosovitskiy, Gupta, Koltun, Kosecka, Malik, Mottaghi, Savva et~al.}]{anderson2018evaluation}
Peter Anderson, Angel Chang, Devendra~Singh Chaplot, Alexey Dosovitskiy, Saurabh Gupta, Vladlen Koltun, Jana Kosecka, Jitendra Malik, Roozbeh Mottaghi, Manolis Savva, et~al. 2018{\natexlab{a}}.
\newblock On evaluation of embodied navigation agents.
\newblock \emph{arXiv preprint arXiv:1807.06757}.

\bibitem[{Anderson et~al.(2018{\natexlab{b}})Anderson, Wu, Teney, Bruce, Johnson, Sünderhauf, Reid, Gould, and van~den Hengel}]{r2r}
Peter Anderson, Qi~Wu, Damien Teney, Jake Bruce, Mark Johnson, Niko Sünderhauf, Ian Reid, Stephen Gould, and Anton van~den Hengel. 2018{\natexlab{b}}.
\newblock Vision-and-language navigation: Interpreting visually-grounded navigation instructions in real environments.
\newblock In \emph{Proceedings of the IEEE Conference on Computer Vision and Pattern Recognition (CVPR)}.

\bibitem[{Bagdasaryan et~al.(2020)Bagdasaryan, Veit, Hua, Estrin, and Shmatikov}]{bagdasaryan2020backdoor}
Eugene Bagdasaryan, Andreas Veit, Yiqing Hua, Deborah Estrin, and Vitaly Shmatikov. 2020.
\newblock How to backdoor federated learning.
\newblock In \emph{International Conference on Artificial Intelligence and Statistics}, pages 2938--2948. PMLR.

\bibitem[{Bernstein et~al.(2018)Bernstein, Zhao, Azizzadenesheli, and Anandkumar}]{bernstein2018signsgd}
Jeremy Bernstein, Jiawei Zhao, Kamyar Azizzadenesheli, and Anima Anandkumar. 2018.
\newblock signsgd with majority vote is communication efficient and fault tolerant.
\newblock \emph{arXiv preprint arXiv:1810.05291}.

\bibitem[{Bhagoji et~al.(2019)Bhagoji, Chakraborty, Mittal, and Calo}]{bhagoji2019analyzing}
Arjun~Nitin Bhagoji, Supriyo Chakraborty, Prateek Mittal, and Seraphin Calo. 2019.
\newblock Analyzing federated learning through an adversarial lens.
\newblock In \emph{International Conference on Machine Learning}, pages 634--643. PMLR.

\bibitem[{Blanchard et~al.(2017)Blanchard, Mhamdi, Guerraoui, and Stainer}]{Blanchard2017MachineLW}
Peva Blanchard, El~Mahdi~El Mhamdi, Rachid Guerraoui, and Julien Stainer. 2017.
\newblock Machine learning with adversaries: Byzantine tolerant gradient descent.
\newblock In \emph{NIPS}.

\bibitem[{Cao et~al.(2021)Cao, Fang, Liu, and Gong}]{Cao2021FLTrustBF}
Xiaoyu Cao, Minghong Fang, Jia Liu, and Neil~Zhenqiang Gong. 2021.
\newblock Fltrust: Byzantine-robust federated learning via trust bootstrapping.
\newblock \emph{ArXiv}, abs/2012.13995.

\bibitem[{Chattopadhyay et~al.(2021)Chattopadhyay, Hoffman, Mottaghi, and Kembhavi}]{chattopadhyay2021robustnav}
Prithvijit Chattopadhyay, Judy Hoffman, Roozbeh Mottaghi, and Aniruddha Kembhavi. 2021.
\newblock Robustnav: Towards benchmarking robustness in embodied navigation.
\newblock In \emph{Proceedings of the IEEE/CVF International Conference on Computer Vision}, pages 15691--15700.

\bibitem[{Chen et~al.(2017)Chen, Liu, Li, Lu, and Song}]{chen2017targeted}
Xinyun Chen, Chang Liu, Bo~Li, Kimberly Lu, and Dawn Song. 2017.
\newblock Targeted backdoor attacks on deep learning systems using data poisoning.
\newblock \emph{arXiv preprint arXiv:1712.05526}.

\bibitem[{Gu et~al.(2017)Gu, Dolan-Gavitt, and Garg}]{Gu2017BadNetsIV}
Tianyu Gu, Brendan Dolan-Gavitt, and Siddharth Garg. 2017.
\newblock Badnets: Identifying vulnerabilities in the machine learning model supply chain.
\newblock \emph{ArXiv}, abs/1708.06733.

\bibitem[{Guo et~al.(2022)Guo, Guo, Wang, and Xu}]{Guo2022PromptFLLF}
Tao Guo, Song Guo, Junxiao Wang, and Wenchao Xu. 2022.
\newblock Promptfl: Let federated participants cooperatively learn prompts instead of models - federated learning in age of foundation model.
\newblock \emph{ArXiv}, abs/2208.11625.

\bibitem[{Hao et~al.(2020)Hao, Li, Li, Carin, and Gao}]{Hao_2020_CVPR_prevalent}
Weituo Hao, Chunyuan Li, Xiujun Li, Lawrence Carin, and Jianfeng Gao. 2020.
\newblock Towards learning a generic agent for vision-and-language navigation via pre-training.
\newblock In \emph{IEEE/CVF Conference on Computer Vision and Pattern Recognition (CVPR)}.

\bibitem[{He et~al.(2022)He, Yang, Feng, Fu, Akula, Jampani, Narayana, Basu, Wang, and Wang}]{he2022cpl}
Xuehai He, Diji Yang, Weixi Feng, Tsu-Jui Fu, Arjun Akula, Varun Jampani, Pradyumna Narayana, Sugato Basu, William~Yang Wang, and Xin~Eric Wang. 2022.
\newblock Cpl: Counterfactual prompt learning for vision and language models.
\newblock \emph{arXiv preprint arXiv:2210.10362}.

\bibitem[{Hong et~al.(2021)Hong, Wu, Qi, Rodriguez-Opazo, and Gould}]{rec-vlnbert}
Yicong Hong, Qi~Wu, Yuankai Qi, Cristian Rodriguez-Opazo, and Stephen Gould. 2021.
\newblock Vln bert: A recurrent vision-and-language bert for navigation.
\newblock In \emph{Proceedings of the IEEE/CVF Conference on Computer Vision and Pattern Recognition (CVPR)}, pages 1643--1653.

\bibitem[{Huang et~al.(2011)Huang, Joseph, Nelson, Rubinstein, and Tygar}]{huang2011adversarial}
Ling Huang, Anthony~D Joseph, Blaine Nelson, Benjamin~IP Rubinstein, and J~Doug Tygar. 2011.
\newblock Adversarial machine learning.
\newblock In \emph{Proceedings of the 4th ACM workshop on Security and artificial intelligence}, pages 43--58.

\bibitem[{Ivkin et~al.(2019)Ivkin, Rothchild, Ullah, Stoica, Arora et~al.}]{ivkin2019communication}
Nikita Ivkin, Daniel Rothchild, Enayat Ullah, Ion Stoica, Raman Arora, et~al. 2019.
\newblock Communication-efficient distributed sgd with sketching.
\newblock \emph{Advances in Neural Information Processing Systems}, 32.

\bibitem[{Ku et~al.(2020)Ku, Anderson, Patel, Ie, and Baldridge}]{rxr}
Alexander Ku, Peter Anderson, Roma Patel, Eugene Ie, and Jason Baldridge. 2020.
\newblock Room-across-room: Multilingual vision-and-language navigation with dense spatiotemporal grounding.
\newblock In \emph{Proceedings of the 2020 Conference on Empirical Methods in Natural Language Processing (EMNLP)}, pages 4392--4412.

\bibitem[{Lee et~al.(2018)Lee, Chen, Hua, Hu, and He}]{lee2018stacked}
Kuang-Huei Lee, Xi~Chen, Gang Hua, Houdong Hu, and Xiaodong He. 2018.
\newblock Stacked cross attention for image-text matching.
\newblock In \emph{Proceedings of the European conference on computer vision (ECCV)}, pages 201--216.

\bibitem[{Li et~al.(2020)Li, Yin, Li, Zhang, Hu, Zhang, Wang, Hu, Dong, Wei, Choi, and Gao}]{oscar}
Xiujun Li, Xi~Yin, Chunyuan Li, Pengchuan Zhang, Xiaowei Hu, Lei Zhang, Lijuan Wang, Houdong Hu, Li~Dong, Furu Wei, Yejin Choi, and Jianfeng Gao. 2020.
\newblock Oscar: Object-semantics aligned pre-training for vision-language tasks.
\newblock In \emph{Computer Vision - {ECCV} 2020 - 16th European Conference, Glasgow, UK, August 23-28, 2020, Proceedings, Part {XXX}}, volume 12375 of \emph{Lecture Notes in Computer Science}, pages 121--137. Springer.

\bibitem[{Liu et~al.(2020)Liu, Huang, Liu, Xu, Ma, Chen, Maybank, and Tao}]{liu2020spatiotemporal}
Aishan Liu, Tairan Huang, Xianglong Liu, Yitao Xu, Yuqing Ma, Xinyun Chen, Stephen~J Maybank, and Dacheng Tao. 2020.
\newblock Spatiotemporal attacks for embodied agents.
\newblock In \emph{European Conference on Computer Vision}, pages 122--138. Springer.

\bibitem[{Liu et~al.(2021{\natexlab{a}})Liu, Yuan, Fu, Jiang, Hayashi, and Neubig}]{liu2021pre}
Pengfei Liu, Weizhe Yuan, Jinlan Fu, Zhengbao Jiang, Hiroaki Hayashi, and Graham Neubig. 2021{\natexlab{a}}.
\newblock Pre-train, prompt, and predict: A systematic survey of prompting methods in natural language processing.
\newblock \emph{arXiv preprint arXiv:2107.13586}.

\bibitem[{Liu et~al.(2021{\natexlab{b}})Liu, Zheng, Du, Ding, Qian, Yang, and Tang}]{liu2021gpt}
Xiao Liu, Yanan Zheng, Zhengxiao Du, Ming Ding, Yujie Qian, Zhilin Yang, and Jie Tang. 2021{\natexlab{b}}.
\newblock Gpt understands, too.
\newblock \emph{arXiv:2103.10385}.

\bibitem[{Lyu et~al.(2020)Lyu, Yu, Ma, Sun, Zhao, Yang, and Yu}]{Lyu2020PrivacyAR}
L.~Lyu, Han Yu, Xingjun Ma, Lichao Sun, Jun Zhao, Qiang Yang, and Philip~S. Yu. 2020.
\newblock Privacy and robustness in federated learning: Attacks and defenses.
\newblock \emph{ArXiv}, abs/2012.06337.

\bibitem[{Mhamdi et~al.(2018)Mhamdi, Guerraoui, and Rouault}]{Mhamdi2018TheHV}
El~Mahdi~El Mhamdi, Rachid Guerraoui, and S{\'e}bastien Rouault. 2018.
\newblock The hidden vulnerability of distributed learning in byzantium.
\newblock In \emph{ICML}.

\bibitem[{Mnih et~al.(2016)Mnih, Badia, Mirza, Graves, Lillicrap, Harley, Silver, and Kavukcuoglu}]{mnih2016asynchronous}
Volodymyr Mnih, Adria~Puigdomenech Badia, Mehdi Mirza, Alex Graves, Timothy Lillicrap, Tim Harley, David Silver, and Koray Kavukcuoglu. 2016.
\newblock Asynchronous methods for deep reinforcement learning.
\newblock In \emph{International conference on machine learning}, pages 1928--1937. PMLR.

\bibitem[{Qi et~al.(2020)Qi, Wu, Anderson, Wang, Wang, Shen, and van~den Hengel}]{reverie}
Yuankai Qi, Qi~Wu, Peter Anderson, Xin Wang, William~Yang Wang, Chunhua Shen, and Anton van~den Hengel. 2020.
\newblock {REVERIE:} remote embodied visual referring expression in real indoor environments.
\newblock In \emph{2020 {IEEE/CVF} Conference on Computer Vision and Pattern Recognition, {CVPR} 2020, Seattle, WA, USA, June 13-19, 2020}, pages 9979--9988. Computer Vision Foundation / {IEEE}.

\bibitem[{Radford et~al.(2021)Radford, Kim, Hallacy, Ramesh, Goh, Agarwal, Sastry, Askell, Mishkin, Clark, Krueger, and Sutskever}]{Radford2021LearningTV}
Alec Radford, Jong~Wook Kim, Chris Hallacy, Aditya Ramesh, Gabriel Goh, Sandhini Agarwal, Girish Sastry, Amanda Askell, Pamela Mishkin, Jack Clark, Gretchen Krueger, and Ilya Sutskever. 2021.
\newblock Learning transferable visual models from natural language supervision.
\newblock In \emph{ICML}.

\bibitem[{Schick and Sch{\"u}tze(2020)}]{schick2020exploiting}
Timo Schick and Hinrich Sch{\"u}tze. 2020.
\newblock Exploiting cloze questions for few shot text classification and natural language inference.
\newblock \emph{arXiv preprint arXiv:2001.07676}.

\bibitem[{Shen et~al.(2022)Shen, Li, Tan, Bansal, Rohrbach, Chang, Yao, and Keutzer}]{shen2022how}
Sheng Shen, Liunian~Harold Li, Hao Tan, Mohit Bansal, Anna Rohrbach, Kai-Wei Chang, Zhewei Yao, and Kurt Keutzer. 2022.
\newblock \href {https://openreview.net/forum?id=zf_Ll3HZWgy} {How much can {CLIP} benefit vision-and-language tasks?}
\newblock In \emph{International Conference on Learning Representations}.

\bibitem[{Tan et~al.(2019)Tan, Yu, and Bansal}]{tan2019learning}
Hao Tan, Licheng Yu, and Mohit Bansal. 2019.
\newblock Learning to navigate unseen environments: Back translation with environmental dropout.
\newblock \emph{arXiv preprint arXiv:1904.04195}.

\bibitem[{Vanhaesebrouck et~al.(2017)Vanhaesebrouck, Bellet, and Tommasi}]{fedavg}
Paul Vanhaesebrouck, Aur{\'{e}}lien Bellet, and Marc Tommasi. 2017.
\newblock Decentralized collaborative learning of personalized models over networks.
\newblock In \emph{Proceedings of the 20th International Conference on Artificial Intelligence and Statistics, {AISTATS} 2017, 20-22 April 2017, Fort Lauderdale, FL, {USA}}, Proceedings of Machine Learning Research, pages 509--517.

\bibitem[{Wang et~al.(2019)Wang, Huang, Celikyilmaz, Gao, Shen, Wang, Wang, and Zhang}]{wang2019reinforced}
Xin Wang, Qiuyuan Huang, Asli Celikyilmaz, Jianfeng Gao, Dinghan Shen, Yuan-Fang Wang, William~Yang Wang, and Lei Zhang. 2019.
\newblock Reinforced cross-modal matching and self-supervised imitation learning for vision-language navigation.
\newblock In \emph{Proceedings of the IEEE/CVF Conference on Computer Vision and Pattern Recognition}, pages 6629--6638.

\bibitem[{Xie et~al.(2019)Xie, Huang, Chen, and Li}]{xie2019dba}
Chulin Xie, Keli Huang, Pin-Yu Chen, and Bo~Li. 2019.
\newblock Dba: Distributed backdoor attacks against federated learning.
\newblock In \emph{International Conference on Learning Representations}.

\bibitem[{Yin et~al.(2018)Yin, Chen, Ramchandran, and Bartlett}]{Yin2018ByzantineRobustDL}
Dong Yin, Yudong Chen, Kannan Ramchandran, and Peter~L. Bartlett. 2018.
\newblock Byzantine-robust distributed learning: Towards optimal statistical rates.
\newblock \emph{ArXiv}, abs/1803.01498.

\bibitem[{Zhang et~al.(2022)Zhang, Panda, Song, Yang, Mahoney, Mittal, Kannan, and Gonzalez}]{zhang2022neurotoxin}
Zhengming Zhang, Ashwinee Panda, Linyue Song, Yaoqing Yang, Michael Mahoney, Prateek Mittal, Ramchandran Kannan, and Joseph Gonzalez. 2022.
\newblock Neurotoxin: Durable backdoors in federated learning.
\newblock In \emph{International Conference on Machine Learning}, pages 26429--26446. PMLR.

\bibitem[{Zhou and Wang(2022)}]{zhou2022fedvln}
Kaiwen Zhou and Xin~Eric Wang. 2022.
\newblock Fedvln: Privacy-preserving federated vision-and-language navigation.
\newblock \emph{arXiv preprint arXiv:2203.14936}.

\bibitem[{Zhou et~al.(2022)Zhou, Yang, Loy, and Liu}]{zhou2022conditional}
Kaiyang Zhou, Jingkang Yang, Chen~Change Loy, and Ziwei Liu. 2022.
\newblock Conditional prompt learning for vision-language models.
\newblock In \emph{Proceedings of the IEEE/CVF Conference on Computer Vision and Pattern Recognition}, pages 16816--16825.

\end{thebibliography}

\clearpage
\appendix

\section{Algorithm Details}
\label{sec:algorithm}
\begin{algorithm} [t]
    \caption{Federated learning with prompt-based aggregation}
    \label{alg:fl}
    \begin{algorithmic}[1]
        \REQUIRE Parameters: participation rate $r$; number of clients $n$; local learning rate $\lambda$; server learning rate $\eta$; number of communication rounds $T$; local training epochs $\tau$. 
        \FOR{$t = 1 \rightarrow T$}
            \STATE Server samples $\lceil rn \rceil$ clients as $\phi_{t}$\
            \STATE Server sends global model and prompts to selected clients $\phi_t$
            \FOR{client $c_i$ in $\phi_{t}$}
                \STATE Client $c_i$ initialization: $(w_{i}^{t-1}, p_{V,i}, p_{L,i}) = (w^{t-1},p_{V,g},p_{L,g})$
                \STATE Client $c_i$ local training: $w_{i}^{t}, p'_{V,i}, p'_{L,i} = {\rm ClientUpdate}(w_{i}^{t-1},  p_{V,i}, p_{L,i}, \tau, \lambda)$
                \STATE Client $c_i$ uploads delta of the language encoder $\Delta w_{i}^{t}=w_{i}^{t}-w^{t-1},\Delta p_{V,i}=p'_{V,i}-p_{V,i},\Delta p_{L,i}=p'_{L,i}-p_{L,i}$ to the server 
            \ENDFOR
            \STATE Server aggregation: $w^t=PBA(\phi_t,\Delta w_{i}^t,\Delta p_{V,i},\Delta p_{L,i},rm)$
        \ENDFOR
    \end{algorithmic}
\end{algorithm}
In prompt-based aggregation (PBA), the visual prompt and the text prompt are learnable vectors. Global visual prompt $p_{V,g}$ or the visual prompt of client $i$ $p_{V,i}$ has the same dimension as the hidden state $h_t$ output from the view encoder, and global text prompt $p_{L,g}$ or the text prompt of client $i$ $p_{L,i}$ has the same dimension as the embedding of each text token $u_1,u_2,u_3,...,u_L$.

When applying PBA in federated learning, at the start of each communication round, both local model weight and local prompts are initialized by global model weight and global prompts. After both local model weight and local prompt parameters are updated through the local training process of each client, we utilize the update of prompt parameters to select some clients to do the aggregation. The whole training procedure is shown in Alg.~\ref{alg:fl}. It's worth noting that only model weight is updated in aggregation, while the global prompts $p_{V,g}$ and $p_{L,g}$ are fixed. 

For the calculation of similarity, the similarity $Sim(i,j)$ between client $i$ and client $j$ is calculated as below:
\begin{equation}
\label{sim}
    \begin{aligned}
    Sim(i,j)=\cos<&Sign([\Delta p_{V,i}, \Delta p_{L,i}]), \\
    &Sign([\Delta p_{V,j}, \Delta p_{L,j}])> 
    \end{aligned}
\end{equation}
where $\Delta p_{V,i}$ and $\Delta p_{L,i}$ are the update of prompt parameters of $i_{th}$ client. We employ the $Sign$ function here as the direction of parameters update is more important than the magnitude in federated learning. For the selection of clients, We apply the similar selection rule in MultiKrum \cite{Blanchard2017MachineLW}, which selects clients with high similarity to others. The detailed procedure of PBA is as shown Alg.~\ref{alg:pba}. 

\begin{algorithm} [t]
    \renewcommand{\algorithmicrequire}{\textbf{Input:}}
	\renewcommand{\algorithmicensure}{\textbf{Output:}}
    \caption{Prompt-based Aggregation (PBA) in communication round $t$}
    \label{alg:pba}
    \begin{algorithmic}[1]
        \REQUIRE the set of sampled clients for this round $\phi_t$; update of model weight of each client $\Delta w_i^{t}$; update of prompt parameters of each client $c_i$ $\Delta p_{V,i}, \Delta p_{L,i}$; expected number of malicious clients $m_e$\\
        \ENSURE global model weight after the aggregation of this round $w^{t}$\\
        \STATE Calculate the similarity $Sim(i,j)$ between each pair of clients in $\phi_t$
        \STATE $S_{rem}=\phi_t$,$S_{sel}=\{\}$
        \WHILE{$|S_{rem}| > 2m_e+2$}
            \FOR{client $c_i$ in $S_{rem}$}
                \STATE Select $|S_{rem}-m_e-1|$ largest $Sim$ values for $c_i$ with other clients in $S_{rem}$, which can be assumed to be $\{Sim(i,1),Sim(i,2),...,Sim(i,|S_{rem}-m_c-1|)\}$ with no harm.
                \STATE Calculate prompt score of $c_i$: $Score(i)=\sum_{j=1}^{|S_{rem}|-m_c-1}Sim(i,j)$
                \ENDFOR
            \STATE Select the client $c_h$ with largest value of prompt score: $c_h=\mathop{\mathrm{argmax}}\limits_{c_i\in S_{rem}}{Score(i)}$
            \STATE Update $S_{rem}$ and $S_{sel}$: $S_{rem}=S_{rem}-c_h,S_{sel}=S_{sel}+c_h$
            \ENDWHILE
        \STATE Aggregation: $w^{t} = w^{t-1} + \eta \sum_{i\in S_{sel}}\frac{n_{j}}{\sum_{j\in S_{sel}}n_{j}} \Delta w_{i}^{t}$
        \STATE \textbf{return} $w^t$
    \end{algorithmic}
\end{algorithm}

In the variant PBA-Input, we use the concatenation of parameters update of visual prompt and text prompt in input position to replace the concatenation of original prompt embeddings in Equ. \ref{sim}. In the variant PBA-Param, we use the parameters of the attention layer to replace the original prompt embeddings in Equ. \ref{sim}. The remaining two variants are the same as PBA.

\section{Experiment Setup}
\label{sec:exp_set}
\vpara{Datasets} \textbf{R2R}~\cite{r2r} uses the Matterport3D region annotations to sample the start and end point pairs, then calculate the shortest paths between them to generate navigation data. 
The dataset contains 7,189 paths from 90 environments.
The environments are split into 61 environments for training and seen validation, 11 for unseen validation, and 18 for testing. 
\textbf{RxR}~\cite{rxr}is proposed to mitigate shortcomings of former VLN datasets. It contains 16,522 paths and 126,069 instructions. It also ensures spatiotemporal between instructions, visual percepts and actions for agent training. 

\vpara{Defense Baselines} Brief descriptions of our defense baselines are given as follows (here we add two more baselines):

\setlist{nolistsep}
    \begin{itemize}[noitemsep, leftmargin=*]
        \item \emph{FedAvg}~\cite{fedavg} is the basic FL aggregation rule.
        \item \emph{Median}~\cite{Yin2018ByzantineRobustDL} aggregates the gradient from clients by calculating the median value of each dimension of the gradients.
        \item \emph{Trimmed Mean}~\cite{Yin2018ByzantineRobustDL} sorts the values of this dimension of all gradients and deletes $m$ maximum and minimum, calculating the average of the remaining values as the aggregation of this dimension.
        \item \emph{Multi-Krum}~\cite{Blanchard2017MachineLW} adopts Krum to select the gradient from the remaining set (initialized as the set of all gradients) and adds it to the selection set (initialized as an empty set), then deletes the selected one from the remaining set. 
        \item \emph{Bulyan} \cite{Mhamdi2018TheHV} adopts Multi-Krum to select gradients, and uses Trimmed Mean to calculate the final gradients.
        \item \emph{FLTrust} \cite{Cao2021FLTrustBF} requires the server has a clean root dataset to approximate the benign gradients.
    \end{itemize}

\vpara{Implementation Details} In datasets, the environments are split into $61$ environments for training and seen validation, $11$ for unseen validation. When training on seen environments, the total number of training steps of local models is the same as centralized training steps. At each communication round, we use the participation rate of $r=0.2$, which indicates that we sample $12$ clients out of $61$ clients for the training of this round. We train each local agent for $\tau=5$ epochs on local data. We set the global learning rate $\eta=2$ following \cite{zhou2022fedvln}. 

During the local training process, the model is trained on its own local dataset for a few epochs under the setting of federated learning, following a hybrid approach termed mixed learning (IL + RL). The local model undertakes separate IL and RL stages, each involving loss computations. The specifics of these IL and RL stages, including the process of loss calculation, are detailed in Sec. \ref{3.2}. 

For the attack, data poisoning is executed by inserting the trigger into the view (see Sec. \ref{3.2}). Each view of a viewpoint is an RGB image, and each viewpoint has multiple views. The trigger we use is a random pattern that is the same size as the image of the view. We implant the trigger into the view by directly summing it and the original RGB image, then generate a new corrupted view. Therefore, changes are made to the environment by modifying its views. The number of malicious clients $m$ is $5$, which indicates that one of the 12 clients in each communication round is malicious in expectation. When applying backdoor attacks during training in malicious clients, the probability of inserting the trigger at each time step $p$ is 0.3, which approximates the fraction of poisoned data. The fix rate $p_m$ is $0.3$. These settings are the default if not mentioned.

The global model at the server is evaluated on seen and unseen validation environments after each communication round. Evaluation metrics except for attack success rate (ASR) are evaluated on clean seen and unseen validation environments. When evaluating ASR, we poison the validation environments with $p=0.1$ and the same trigger utilized by malicious clients during local training. We then calculate ASR by validating the poisoned seen and unseen validation environments.

\section{More Experiment Results}
\label{sec:att_def}
\subsection{Attack}

\begin{figure}
    \centering
    \includegraphics[width=\linewidth]{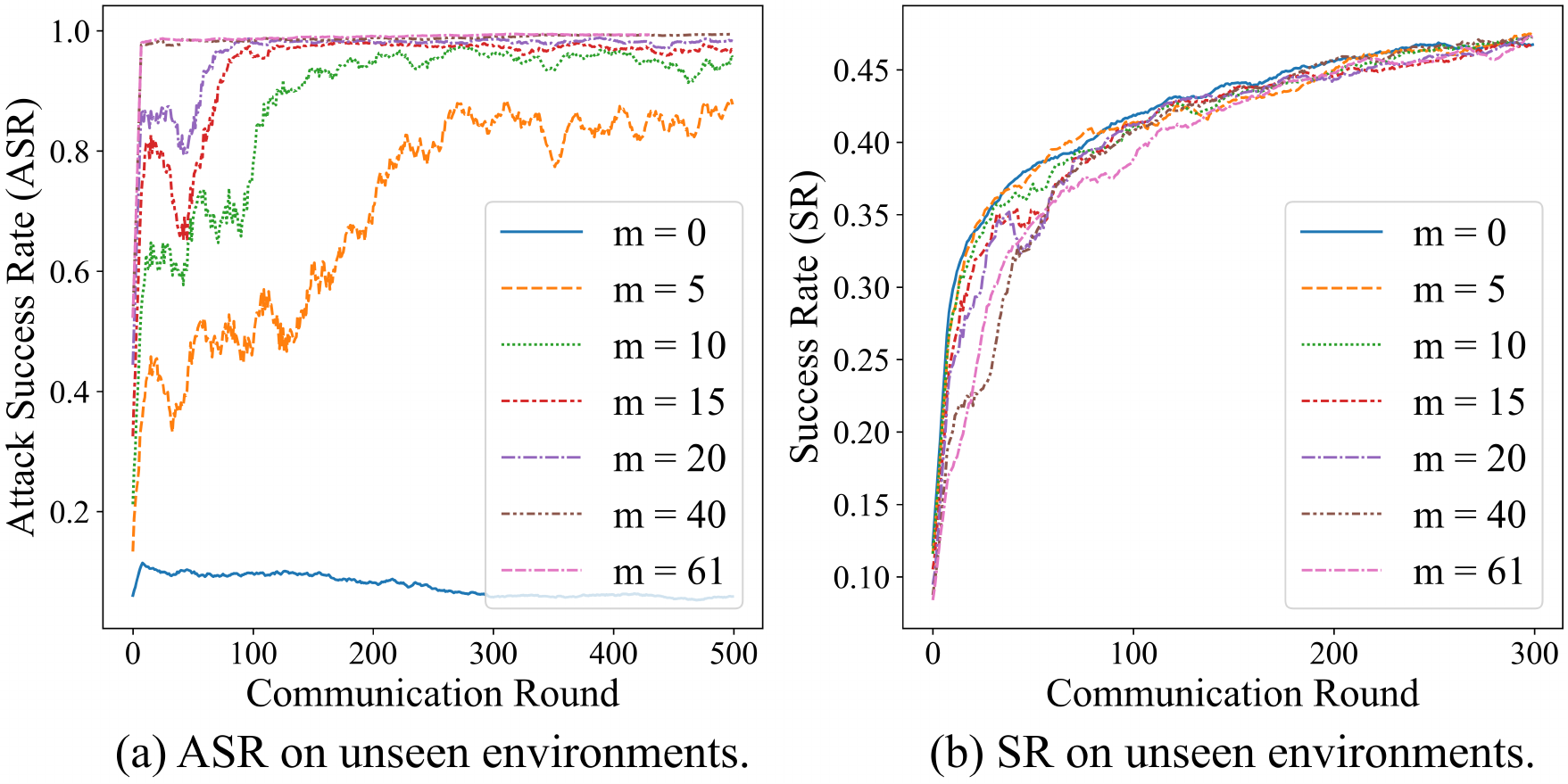}
    \caption{Impact of the number of malicious clients. Results are evaluated on R2R with CLIP-ViL.}
    \label{fig:attack_num}
    \vspace{-1ex}
\end{figure}

\begin{figure}
    \centering
    \includegraphics[width=\linewidth]{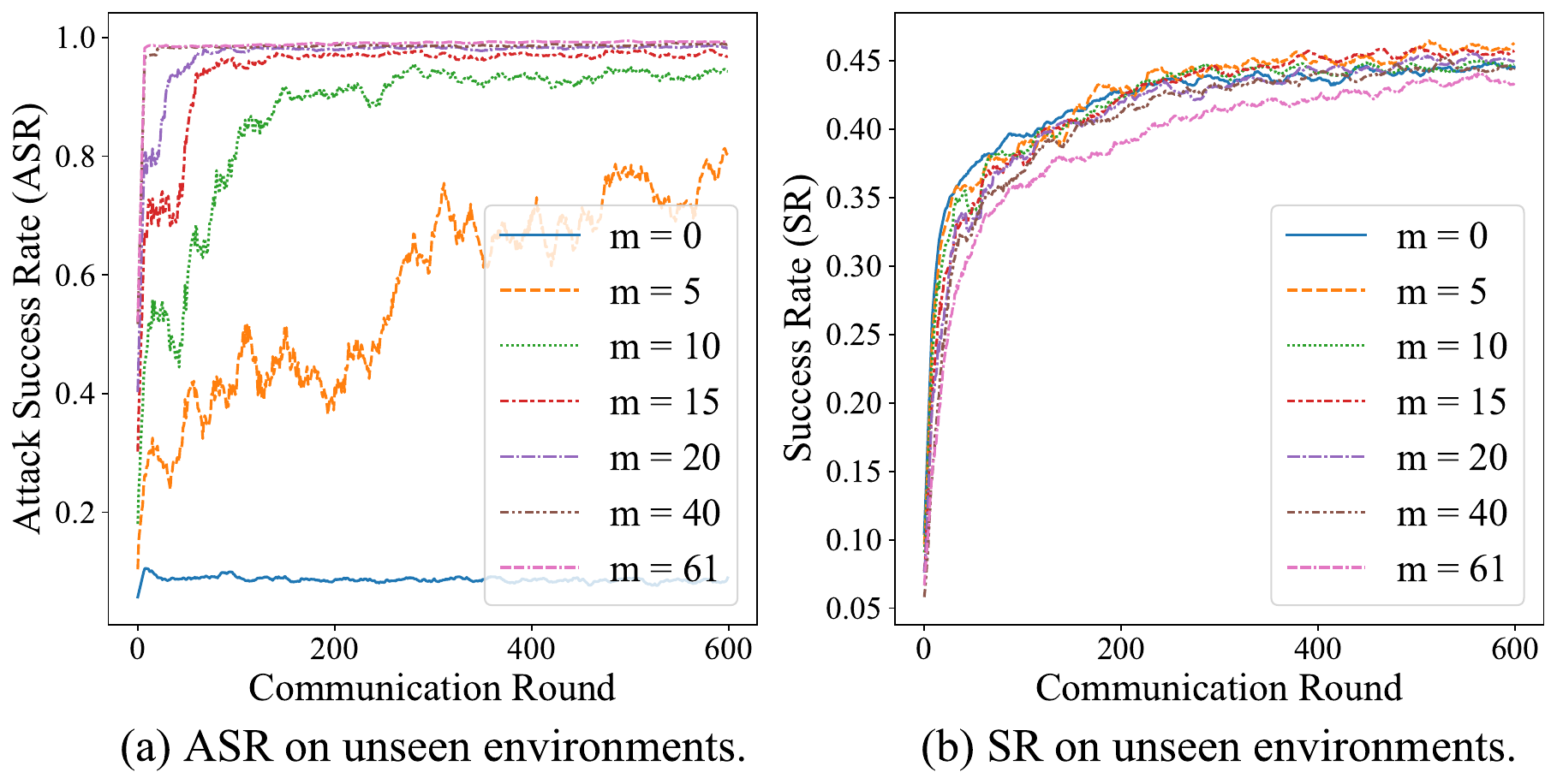}
    \caption{Impact of the number of malicious clients. Results are evaluated on R2R with EnvDrop.}
    \label{fig:att_num_env}
\end{figure}

\vpara{Impact of the number of malicious clients.} 
Fig. \ref{fig:attack_num} shows the results under different numbers of malicious clients with CLIP-ViL. In Fig.~\ref{fig:attack_num}(a), we can observe that the number of malicious clients is positively correlated with ASR and its increase accelerates the convergence of ASR. 
For SR in Fig.~\ref{fig:attack_num}(b),  more malicious clients would cause a greater fluctuation of SR during the first 100 communication rounds.
Therefore, comparing the results with that of $m=0$, the attack under $m \ge 20$ cannot achieve the backdoor attack requirement. 

Fig. \ref{fig:att_num_env} shows the results under different numbers of malicious clients with EnvDrop. In Fig.~\ref{fig:att_num_env}(a), we can observe that the increase in the number of malicious clients not only accelerates the convergence of ASR, but also improves the final ASR. 
For SR in Fig.~\ref{fig:att_num_env}(b),  more malicious clients would cause an obvious performance drop during training. Comparing the results with that of $m=0$, we can find that the attack under $m \ge 20$ cannot achieve the expected backdoor attack goal, which requires the performance of the attacked model on the clean dataset to keep the same level as that of the unattacked model.

\begin{figure}
    \centering
    \includegraphics[width=\linewidth]{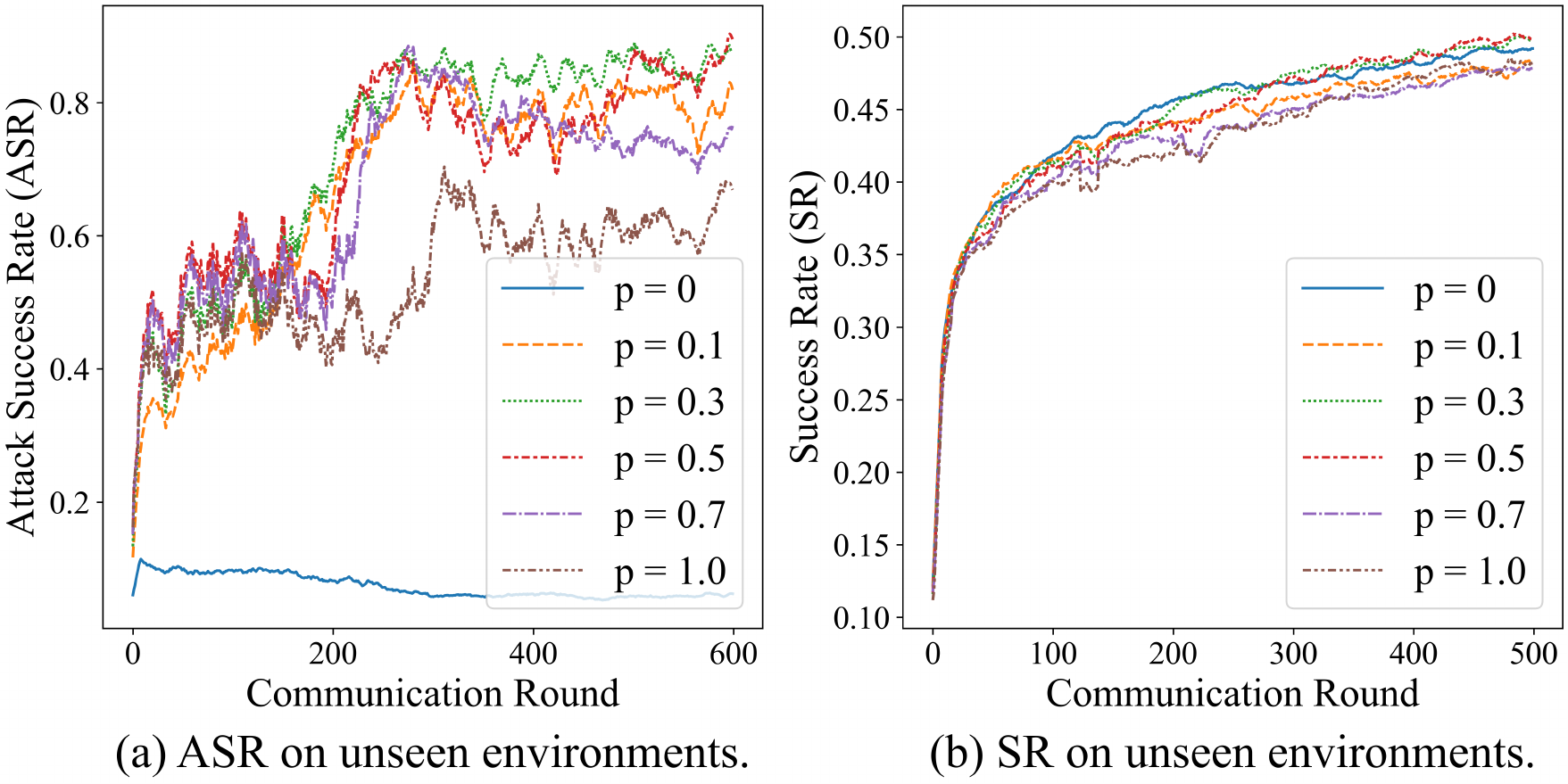}
    \caption{Impact of the fraction of poisoned data. Results are evaluated on unseen environments in R2R with CLIP-ViL.}
    \label{fig:attack_frac}
    \vspace{-2ex}
\end{figure}

\begin{figure}
    \centering
    \includegraphics[width=\linewidth]{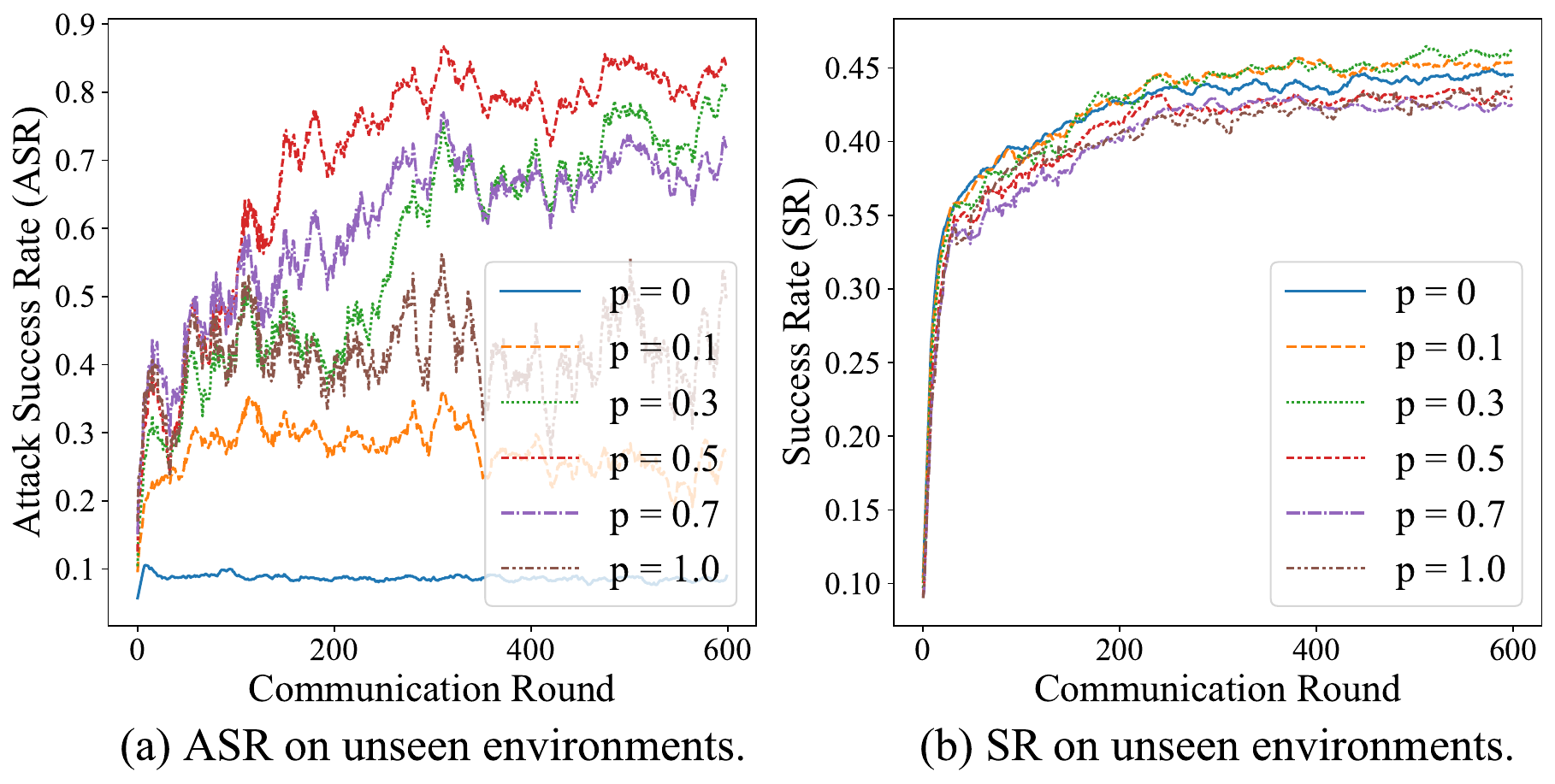}
    \caption{Impact of the fraction of poisoned data. Results are evaluated on R2R with EnvDrop.}
    \label{fig:att_frac_env}
\end{figure}

\vpara{Impact of the fraction of poisoned data.} 
$p$ approximates the fraction of poisoned data during training. Fig.~\ref{fig:attack_frac} shows both SR and ASR of FedVLN agents under different fractions of poisoned data with CLIP-ViL. For ASR, a larger fraction does not lead to a higher ASR; on the contrary, it obtains an even lower ASR than a smaller fraction of poisoned data. 
One possible reason is that a large fraction of poisoned data affects the scene understanding of the agent, making it harder to recognize the trigger. SR becomes lower when the fraction of poisoned data is higher. When $p \ge 0.5$, the drop in performance is obvious, inducing a nearly 3\% SR gap. This result indicates that there is no strong correlation between ASR and the fraction.

Fig.~\ref{fig:att_frac_env} shows both SR and ASR of FedVLN agents under different fractions of poisoned data with EnvDrop. For ASR, we can find that a larger fraction of poisoned data could not lead to a better attack. ASR of $p=0.1$ and $p=1.0$ are quite close.
SR becomes lower when the fraction of poisoned data is higher, while ASR of $p=0.3$ and $p=0.5$ are high. It indicates that we need to select an appropriate range for $p$ to achieve great attack effects. For SR, When $p \ge 0.5$, the drop in performance is obvious, inducing a nearly 6\% SR gap. It proves that a larger fraction of poisoned data could hurt the performance of the attacked model on the clean dataset, which is not expected in the backdoor attack.

\begin{figure}[htp!]
    \centering
    \includegraphics[width=\linewidth]{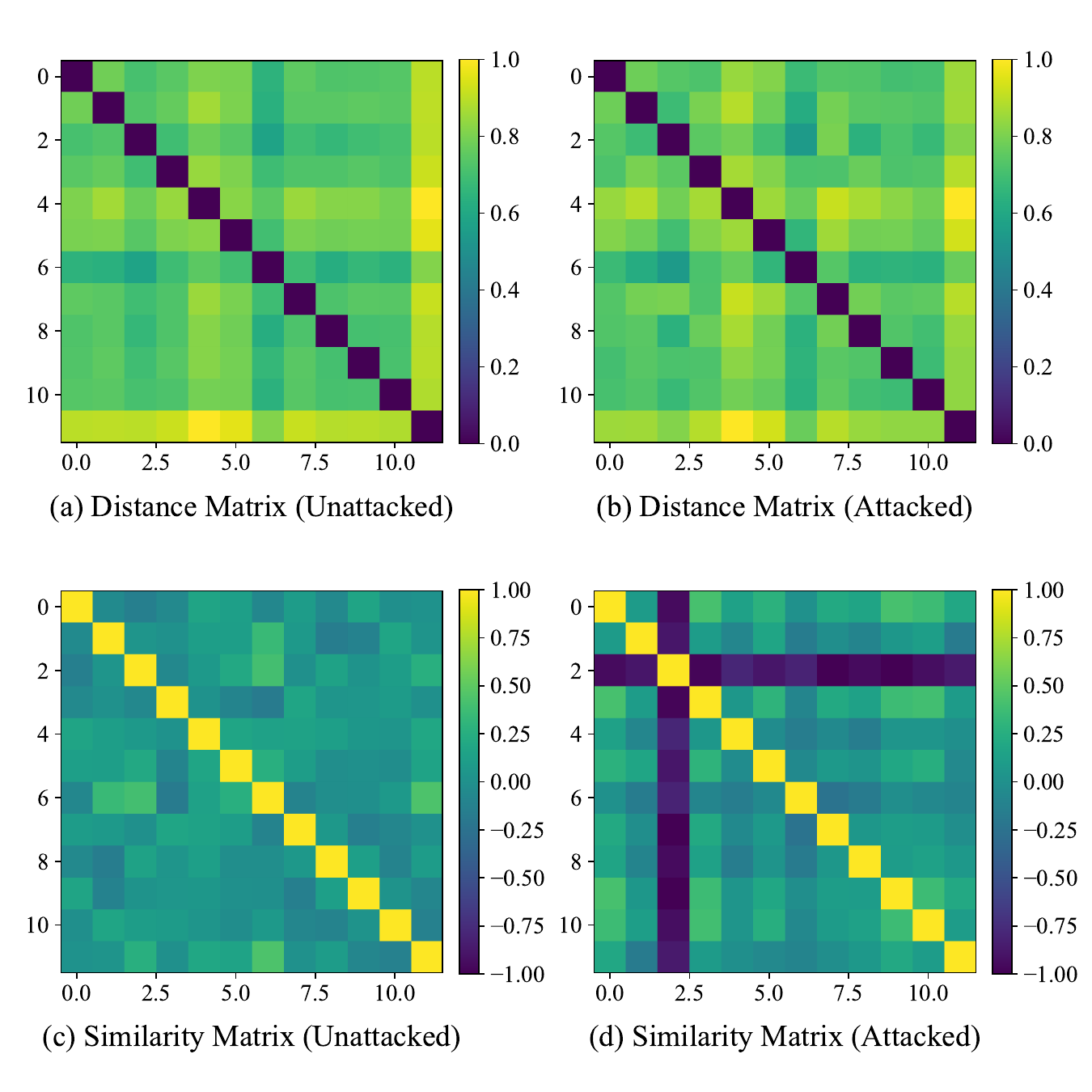}
    \caption{The illustration of the difference of the method to calculate the distance matrix and similarity matrix of PBA-Param ((a) and (b)) and PBA ((c) and (d)). All the matrix is $12\times12$ because there are 12 clients in every round. The matrix represents the distance (in PBA-Param) or the similarity (in PBA) of the distribution of specific part of parameters between the 12 clients. For the specific part of parameters, it would be the attention layer in PBA-Param and the prompt in PBA. The diagonal of the distance matrix is 0 and the similarity matrix is 1. }
    \label{fig:mat}
\end{figure}

\subsection{Defense}
\begin{figure}
    \centering
    \includegraphics[width=0.9\linewidth]{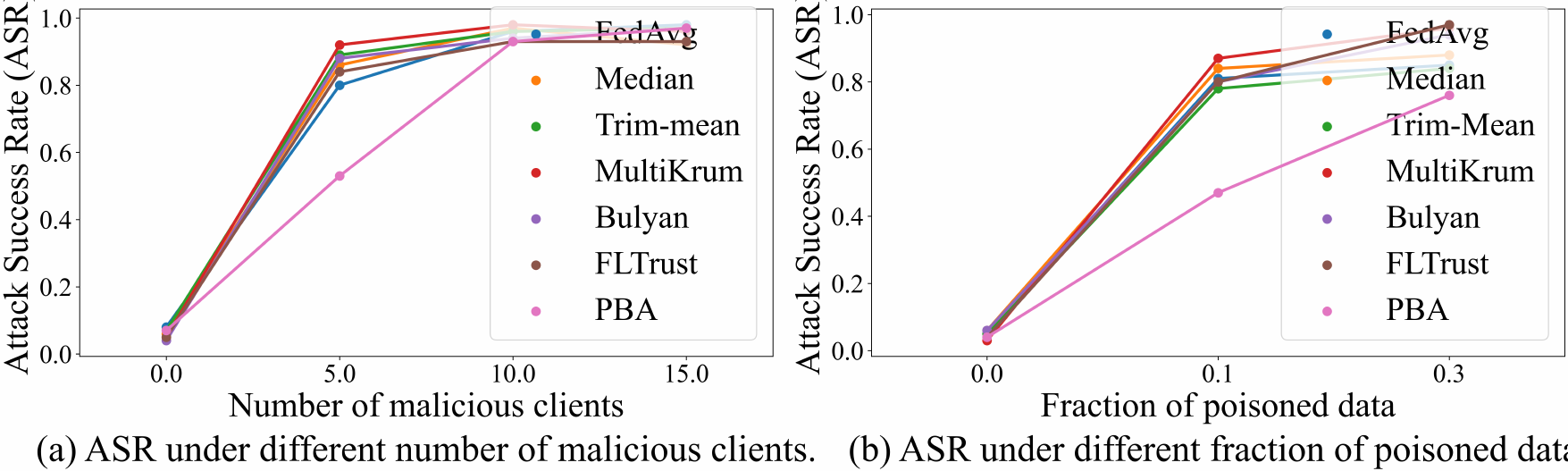}
    \vspace{0.5ex}
    \caption{Impact of the number of malicious clients and the fraction of poisoned data in the unseen environments with CLIP-ViL.}
    \label{fig:def}
    \vspace{-2ex}
\end{figure}

\vpara{Impact of the fraction of poisoned data and the number of malicious clients.}  In Fig. \ref{fig:def}, we only visualize the results where values of these two factors successfully meet the requirement of the backdoor attack.
Fig.~\ref{fig:def}(b) shows that \defense significantly outperforms any other defense methods on different fractions of poisoned data. For the number of malicious clients, ASR of different defense methods is nearly 100\% when there are too many malicious clients (\eg, $m \ge 10$). \defense still outperforms other defense methods when $m \le 10$. 

\begin{figure}
    \centering
    \includegraphics[width=\linewidth]{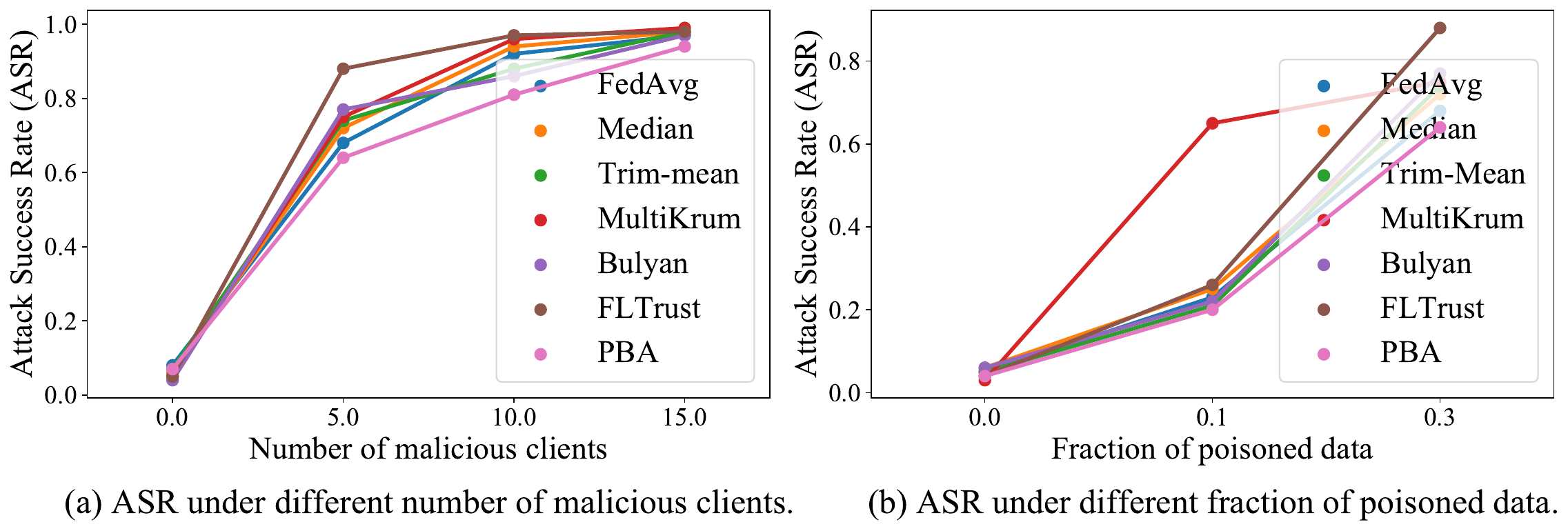}
    \caption{Impact of the number of malicious clients. Results are evaluated on unseen environments of R2R with EnvDrop.}
    \label{fig:def_env}
\end{figure}
\vpara{Impact of the number of malicious clients and fraction of poisoned data.} Fig. \ref{fig:def_env} shows the results of different defense methods under different fractions of poisoned data and different numbers of malicious clients with EnvDrop. For the number of malicious clients $m$, ASR of different defense methods are close to 100\% when there are too many malicious clients (\eg, $m \ge 10$).
For the fraction of poisoned data, it is shown in Fig.~\ref{fig:def_env}(b) that ASR of different defense methods mostly maintains the same level as that of FedAvg. Some methods (\eg, MultiKrum) even exacerbate it. For instance, when $p=0.1$, ASR of MultiKrum is almost three times that of FedAvg. On the whole, PBA significantly outperforms any other defense methods in each case. 

\subsection{Adaptive Adversary}

\begin{table}[!t]
\resizebox{\columnwidth}{!}{
\begin{tabular}{cccccc} 
\toprule
\multirow{2}*{Attack} & \multirow{2}*{Method} & \multicolumn{2}{c}{Val-Seen} & \multicolumn{2}{c}{Val-Unseen}  \\ \cmidrule(lr){3-4} \cmidrule(lr){5-6}
~ & ~ & EnvDrop & CLIP-ViL & EnvDrop & CLIP-ViL \\
\midrule
    NAW & FedAvg & 0.71 & 0.87 & 0.68 & 0.85 \\
    NAW & Median & 0.70 & 0.89 & 0.72 & 0.88 \\
    NAW & Trim-Mean & 0.76 & 0.86 & 0.74 & 0.84 \\
    NAW & MultiKrum & 0.77 & 0.95 & 0.75 & 0.96 \\
    NAW & Bulyan & 0.78 & 0.91 & 0.77 & 0.94 \\
    NAW & FLTrust & 0.87 & 0.93 & 0.88 & 0.97 \\
    NAW & PBA & \bf{0.63} & \bf{0.72} & \bf{0.64} & \bf{0.76} \\
    NAW2 & PBA & 0.67 & 0.79 & 0.68 & 0.80 \\
\bottomrule
\end{tabular}
}
\vspace{0.5ex}
\caption{
Comparison of Attack Success Rate (ASR) between different defense methods on R2R and RxR. 
Lower is better. Add the adaptive adversary NAW2 to attack against PBA.
}
\label{table:v2}
\vspace{-2ex}
\end{table}

Here we talk about the possible adaptive adversary of our defense method PBA and test its performance against PBA.

As the local client is entirely controlled by the attacker, one possible adaptive adversary of PBA is that attackers can simply return the global-distributed prompts instead of their updated prompts. Thus, the server can not get the prompt from the malicious client that can tell the alignment variance. We test this strategy in Table \ref{table:v2} (we name this adaptive adversary \textit{NAW2}).

Upon examination, it emerges that despite experiencing an increase of 4\% to 9\% in attack success rate when defending against NAW2, PBA still performs better than other state-of-the-art defense techniques. NAW2 does not succeed in bypassing PBA. This outcome is attributed to the updates of local prompts during local training, and learning the alignment. If attackers only return the global distributed prompts which are fixed during the entire federated learning, these prompts would lack the high similarity characteristic of the updated prompts from other benign clients. Consequently, they would be easily distinguishable and filtered out. It's worth noting that these global distributed prompts are initialized at the outset of the federated learning process and maintain a fixed state throughout training. As such, attempting to return these fixed distributed prompts would prove ineffective.

It should be noted that this adaptive adversary is based on the premise that the attacker has full knowledge of the model, which does not satisfy the black-box setting in the application of FedVLN. However, extending PBA to the white-box scenario and figuring out how to improve it is worth exploring.

\subsection{Why PBA Works?}
\label{work}
We choose one of the rounds in our experiment and present the case study to illustrate the differences between PBA and PBA-Param, as shown in Fig. \ref{fig:mat}. We can see that PBA-Param cannot distinguish the malicious client as the distances of distribution of the update of attention layer parameters between different clients are quite fixed, while our methods can detect the malicious client clearly, demonstrating the importance of analyzing a smaller amount of parameters for precision.

\end{document}